\documentclass[10pt,twocolumn,letterpaper]{article}
\usepackage[pagenumbers]{cvpr}            
\usepackage{graphicx}
\usepackage{amsmath}
\usepackage{amssymb}
\usepackage{booktabs}
\usepackage{balance}
\usepackage{lipsum}
\usepackage{caption}
\usepackage{algorithm}
\usepackage{algorithmic}
\usepackage{enumerate}
\usepackage{amsmath,amsthm,amssymb}
\usepackage{engord}
\usepackage{bm}
\usepackage{bbm}
\usepackage{amstext}
\usepackage{comment}
\usepackage{color}
\usepackage[x11names,table]{xcolor}
\usepackage{ctable}
\usepackage{booktabs}
\usepackage{multirow}
\usepackage{mathtools}
\usepackage{setspace}
\usepackage{pifont}
\usepackage{dsfont} 
\usepackage[inline]{enumitem}
\usepackage[toc,page]{appendix}
\usepackage[pagebackref=true,breaklinks=true,letterpaper=true,colorlinks,citecolor=citecolor,bookmarks=true]{hyperref}
\usepackage[capitalize]{cleveref}
\definecolor{Gray}{gray}{0.9}
\definecolor{White}{gray}{1}
\definecolor{DGray}{gray}{0.8}
\definecolor{citecolor}{HTML}{0071bc}
\newcommand{\tabincell}[2]{\begin{tabular}{@{}#1@{}}#2\end{tabular}}
\newcommand{\minitab}[2][l]{\begin{tabular}{#1}#2\end{tabular}}
\DeclareMathAlphabet\mathbfcal{OMS}{cmsy}{b}{n}

\crefname{section}{Sec.}{Secs.}
\Crefname{section}{Section}{Sections}
\Crefname{table}{Table}{Tables}
\crefname{table}{Tab.}{Tabs.}

\def\confName{CVPR}
\def\confYear{2022}
 
\newcommand\oakink{\textit{OakInk}\xspace} 
\newcommand\oak{\textit{Oak}\xspace} 
\newcommand\ink{\textit{Ink}\xspace} 
\newcommand\tink{\textit{Tink}\xspace}
\newcommand\supp{\textbf{Appx}\xspace}
\newcommand\camsys{$\mathbfcal{S}_{c}$\xspace}
\newcommand\capsys{$\mathbfcal{S}_{m}$\xspace}

\newcommand{\xmark}{\ding{55}}%
\newcommand{\greencheck}{{\color{Green4} \checkmark}}
\newcommand{\redcross}{{\color{red} \xmark}}

\newcommand{\customfootnotetext}[2]{{
  \renewcommand{\thefootnote}{#1}
  \footnotetext[0]{#2}}}

\begin{document}

\title{
    OakInk: A Large-scale Knowledge Repository for Understanding \\ Hand-Object Interaction
}
\author{
    {$^{1,2}$Lixin Yang\textsuperscript{$\star$},}
    ~~{$^{1}$Kailin Li\textsuperscript{$\star$},}
    ~~{$^{1}$Xinyu Zhan\textsuperscript{$\star$},}
    ~~{$^{1}$Fei Wu,}
    ~~{$^{1}$Anran Xu,}
    ~~{$^{1}$Liu Liu,}
    ~~{$^{1,2}$Cewu Lu\textsuperscript{$\dagger$}}\\
    {{$^{1}$Shanghai Jiao Tong University, China}
    ~~~
    {$^{2}$Shanghai Qi Zhi Institute, China}}\\
{\tt\small \{{siriusyang}, {kailinli}, {kelvin34501}, {legendary}, {xuanran},  {liuliu1993}, {lucewu}\}@{sjtu.edu.cn}}
\vspace{-3mm}
}
\maketitle

\customfootnotetext{$\star$}{ Equal contribution.}
\customfootnotetext{$\dagger$}{ 
    Cewu Lu is the corresponding author. He is the member of Qing Yuan Research Institute and MoE Key Lab of Artificial Intelligence, AI Institute, Shanghai Jiao Tong University, and Shanghai Qi Zhi institute, China.
}

\begin{abstract}
    Learning how humans manipulate objects requires machines to acquire knowledge from two perspectives: one for understanding object affordances and the other for learning human's interactions based on the affordances. Even though these two knowledge bases are crucial, we find that current databases lack a comprehensive awareness of them. In this work, we propose a multi-modal and rich-annotated knowledge repository, \oakink, for visual and cognitive understanding of hand-object interactions.
    We start to collect 1,800 common household objects and annotate their affordances to construct the first knowledge base: \oak.
    Given the affordance, we record rich human interactions with 100 selected objects in \oak.
    Finally, we transfer the interactions on the 100 recorded objects to their virtual counterparts through a novel method: \tink.
    The recorded and transferred hand-object interactions constitute the second knowledge base: \ink.
    As a result, \oakink contains 50,000 distinct affordance-aware and intent-oriented hand-object interactions.
    We benchmark \oakink on pose estimation and grasp generation tasks.
    Moreover, we propose two practical applications of \oakink: intent-based interaction generation and handover generation.
    Our datasets and source code are publicly available at 
    \url{https://github.com/lixiny/OakInk}.
\end{abstract}

\begin{figure}[t]
    \begin{center}
      \includegraphics[width=1.0\linewidth]{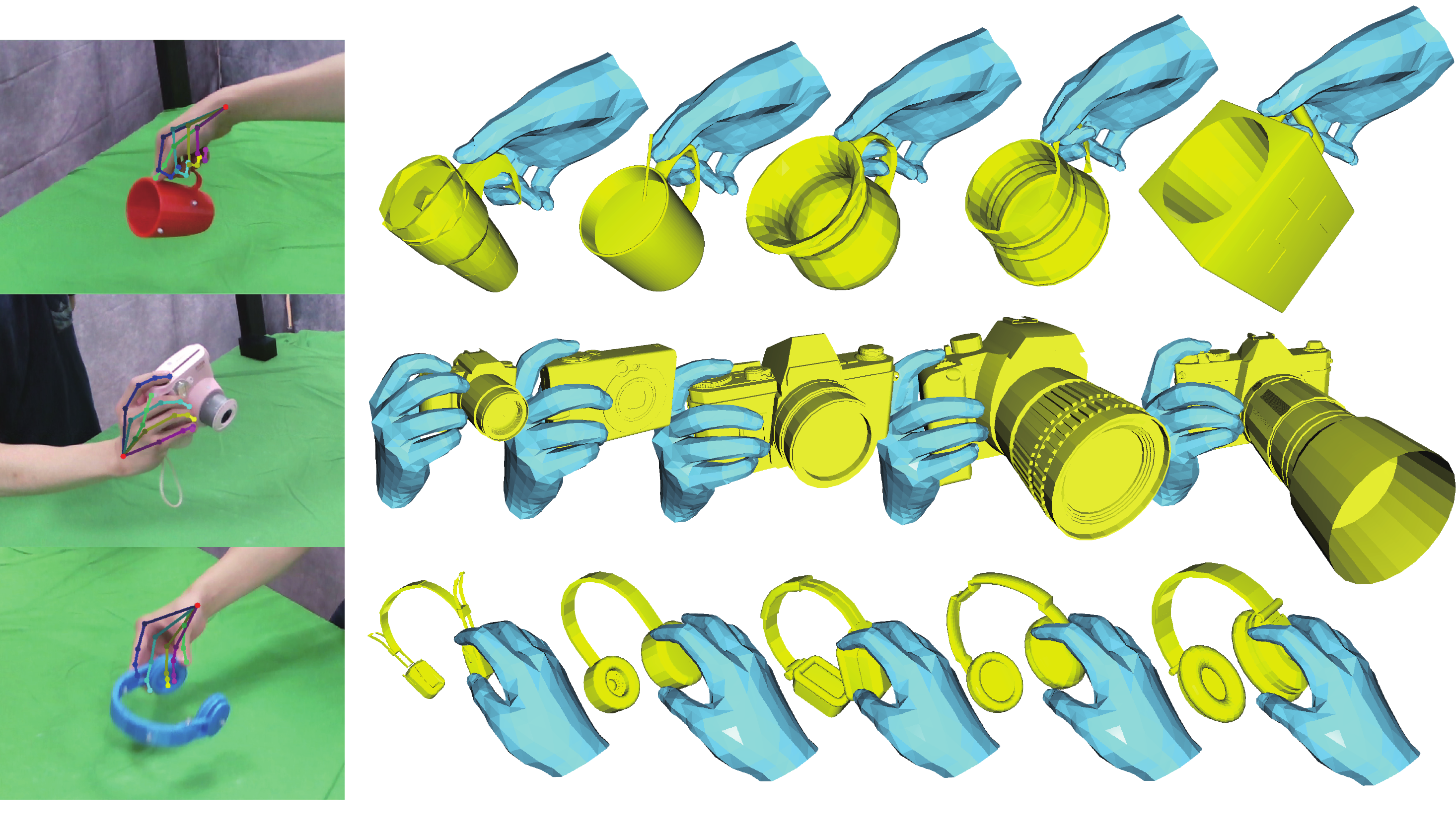}
    \end{center}
    \vspace{-6.5mm}
    \caption{Illustration of different data modalities in \oakink repository. The left column shows human manipulating 3 source objects (mug, camera, and headphones). The right 5 columns show the transferred interactions on 15 virtual counterpart objects. }\vspace{-5.5mm}
    \label{fig:capture}
\end{figure}

\vspace{-5mm}\section{Introduction}
Enabling a machine to understand and imitate the behavior of humans has been
a long-term vision in the history of science.
Among the tasks derived from it,  
learning \textit{how humans manipulate objects} is a fundamental challenging one.
As most tools are designed for achieving function, human can easily learn to manipulate them through instruction or experiences. 
However, these experiences are hard for a machine to acquire. 
It was not until recently that data-driven approaches have begun to promote research on learning human manipulation \cite{Handa2020DexPilot, mo2021where2act, garcia2018fphab, Understanding, Antotsiou2018TaskOrientedHM, Zhu2021thg}.
Prior work has tried to empower a machine complex skills such as hand-object localization \cite{Shan20InternetScale}, pose estimation \cite{li2021artiboost}, grasp generation \cite{corona2020ganhand}, and action imitation \cite{qin2021dexmv}.

\vspace{-0.5mm}Two fundamental components for learning human manipulation are
\textbf{1)} \textit{the affordance of the objects} and 
\textbf{2)} \textit{how human hand would interact with the objects based on those affordances}. 
While the word ``affordance'' has different formulations in different tasks, in this paper, we denote ``affordance'' as the functionality of object.
Since 2019, there have been at least 9 datasets of hand-object interaction released: ObMan\cite{hasson2019obman}, YCBAfford\cite{corona2020ganhand}, HO3D\cite{hampali2020ho3dv2}, ContactPose\cite{brahmbhatt2020contactpose}, GRAB\cite{taheri2020grab}, DexYCB\cite{chao2021dexycb}, two H2O\cite{kwon2021h2o, ye2021h2o} and DexMV\cite{qin2021dexmv}. 
However, these datasets lack comprehensive awareness of the object's affordance and the hand's interactions with it.
First, existing real-world datasets only contain a small number of objects and hand interactions. 
As two illustrative examples, only 20 objects were captured in DexYCB,  
and only 2.3K distinct interactions were captured among 2.9M images in ContactPose (0.08\%). 
Second, even if synthetic dataset \cite{hasson2019obman} can extend to large numbers of interactions in grasp simulator: GraspIt \cite{graspit}, the generated grasps neither reflect the distribution of human interactions nor consider the object's affordance itself.
To understand how humans manipulate objects, we propose to build the machine's knowledge from two perspectives:  object-centric and human-centric perspective.
To this end, we construct two interrelated knowledge bases. 
One is an \textbf{O}bject \textbf{A}ffordance \textbf{K}nowledge base (\textbf{\oak} base, \cref{sec:oakn}) 
    in which we provide comprehensive descriptions of objects' affordances within a knowledge graph, 
and the other is  an \textbf{In}teraction \textbf{K}nowledge base (\textbf{\ink} base, \cref{sec:ikn}) 
    in which we collect diverse human hand interactions that provide demonstrations of manipulating the object according to its affordances.

To construct the \oak base, we firstly collect 1,800 household objects that are designed for single-hand manipulation. 
The sources of objects in \oak base are four-fold: 1) self-collected from online vendors, 2) ShapeNet \cite{shapenet2015} models, 3) YCB\cite{ycb2015} and 4) ContactDB\cite{brahmbhatt2018contactdb} objects.
Second, through exhaustively reviewing the objects in the above sources, 
we build an object knowledge graph that arranges objects with two types of abstractions, namely \textit{taxonomy} and \textit{attribute} (\cref{fig:obj_affod}).
This object knowledge graph enables us to make a quick extension for new objects and conduct convenient clustering for objects of similar affordance.

To construct the \ink base, we start to collect human experiences on performing hand-object interactions based on the object's affordance.
We select 100 representative objects from \oak base, invite 12 human subjects to perform demonstrations, and set up a multi-sensor MoCap platform for recording (\cref{fig:hardware}). 
The recorded sequences constitute a real-world image dataset that contains 230,064 RGB-D frames capturing 12 subjects performing up to 5 intent-oriented hand interactions with objects in a pool of total 100 instances from 32 categories. 
The objects that appeared in the recorded sequences are denoted as the ``\textit{source}'' objects. 
Next, given the real-world human demonstration, we aim to transfer their experience on the \textit{source} object to its virtual counterparts with similar affordances (\textit{target} objects).
The transferred hand interaction should not only ensure its physical plausibility, but also keep the consistent intent and match the size, shape, and affordance of the \textit{target} object (\cref{fig:capture}).
To this end, we propose a learning-fitting hybrid method: \textbf{\tink} for \textbf{T}ransferring the \textbf{In}teraction \textbf{K}nowledge among objects (\cref{sec:transfer}).
\tink consists of three modules: namely an implicit shape interpolation, an explicit contact mapping, and an iterative pose refinement.
With \tink, we extend the total number of \textbf{distinct} hand-object interactions in \ink base to 50,000.

Through combining the above two knowledge bases: \oak and \ink, we construct a large-scale knowledge repository: \textbf{\oakink}.
The advantages of our \oakink are three-fold:
\begin{enumerate*}[label={\arabic*)},font={\bfseries}]
    \item It provides comprehensive knowledge for understanding hand-object interactions from two perspectives: object affordances and human experiences;
    \item It contains two large-scale datasets of image-based and geometry-based  hand-object interaction;
    \item It provides rich annotations including hand and object poses, scanned object models, affordances, fine-grained contact and stress patterns, and intents labels.
\end{enumerate*}
\oakink can benefit researches in two communities:  
\begin{enumerate*}[label={\arabic*)},font={\bfseries}]
    \item pose estimation \cite{hasson2020leveraging, liu2021semi, Doosti2020hopenet}, shape reconstruction \cite{hasson2019obman, karunratanakul2020grasping}, and action recognition \cite{garcia2018fphab, kwon2021h2o, tekin2019h+o} in computer vision, CV;
    \item grasp generation \cite{jiang2021graspTTA, Zhu2021thg, taheri2020grab} and motion synthesis \cite{Cao2020LongtermHM, petrovich21actor} in computer graphics, CG;
\end{enumerate*}
Among all the topics above, we find pose estimation and pose generation are most relevant to our interests.  In this paper, we benchmark \oakink on three existing tasks and propose two new tasks: One is an intent-based hand pose generation and the other is a human-to-human handover generation. 

Our contributions are concluded in three-fold. First,  we construct \oakink, a large-scale knowledge repository for understanding hand-object interactions.
Second, inside \oakink, we propose a novel method \tink that transfers the interaction knowledge among objects with similar affordance. 
Finally, we provide extensive evaluations for benchmarking \oakink on three existing tasks and propose two novel tasks: generating plausible hand poses for more customized purposes.

\vspace{-2mm}\section{Related Work}
\vspace{-1.5mm}\noindent\textbf{Datasets of Hand-Object Interaction (HOI).}
Current HOI datasets can be categorized as real-world and synthetic based on the data source.
ObMan\cite{hasson2019obman} and YCBAfford \cite{corona2020ganhand} represent the synthetic datasets that leveraged grasp simulators to synthesize or label static grasps.
Real-world datasets are categorized into three types based on how they collect the annotations.
\begin{enumerate*}[label={\arabic*).},font={\bfseries}]
    \item The \textit{marker-based} datasets collect hand poses with the aid of hand-attached magnetic sensors \cite{garcia2018fphab,ye2021h2o, Yuan2017BigHand22MBH} or reflective markers \cite{taheri2020grab}.
    \item  The \textit{automatic marker-less} datasets \cite{brahmbhatt2020contactpose, hampali2020ho3dv2} aggregate the visual cues from methods of detection, segmentation and pose estimation to acquire annotations automatically.
    \item The \textit{crowd-sourced marker-less} datasets leverage human annotators to label the 2D poses of hands and objects \cite{chao2021dexycb}.
\end{enumerate*}
In this paper, we collect 3D hand pose annotations through crowd-sourcing its 2D keypoints and optimizing them within multi-views. For object pose, we record its surface-attached reflective wafers in a synchronized MoCap system (\cref{sec:data_anno}).
We will present the comprehensive comparisons and statistics of existing datasets in \cref{sec:analy} (\cref{tab:dataset_comp}).

\vspace{0.5em}\noindent\textbf{Contact of Hand-Object Interaction.}
In order to capture contact, previous methods used measurement devices like force transducers \cite{pham2017hand}, tactile sensors \cite{sundaram2019learning} and thermal cameras \cite{brahmbhatt2018contactdb, brahmbhatt2020contactpose}, or computed realistic contact through accurate pose tracking \cite{taheri2020grab}.
As contact can provide rich cues to reason the conjoint hand-object poses during interactions,
recent methods leveraged contact to help optimize grasps in reconstruction \cite{karunratanakul2020grasping, yang2021cpf} and synthesis \cite{jiang2021graspTTA} tasks.
In this paper, we derive contact regions and their stress patterns through accurate pose tracking (\cref{par:contact}). Later in  \cref{sec:transfer}, we map the contacts to virtual objects and optimize poses based on the contacts.

\vspace{0.5em}\noindent\textbf{Pose of Hand-Object Interaction.}
Pose estimation is a common task for understanding how human manipulate objects.
Previous methods either focus on hand \cite{tzionas2016capturing} or object \cite{tzionas20153d} pose alone.
Hasson \etal \cite{hasson2019obman} have proposed the first conjoint hand-object pose estimation methods that brought the renaissance in this area \cite{hasson2020leveraging, hasson21homan, rhoi2020, liu2021semi, Doosti2020hopenet, li2021artiboost}.
Another popular task of HOI is grasping pose generation. Researchers in this track have delved into synthesizing prehensile grasps based on image \cite{corona2020ganhand} or shape observation \cite{karunratanakul2020grasping}.
Many derivative tasks like: action recognition \cite{garcia2018fphab, kwon2021h2o}, imitation learning \cite{Radosavovic2020}, teleoperation\cite{Handa2020DexPilot}, and human-to-robot handover (and vise versa) \cite{yang2020human, suay2015position} are powered by the above two tasks.
In this paper, we benchmark our dataset on the classical pose estimation and pose generation tasks.
We also introduce two interesting tasks that explore the generative model with a given intent and within a handover scenario.

\vspace{-1mm}\section{Constructing the \oakink}
\vspace{-1.5mm}\oakink consists of two interrelated knowledge bases.
One is the object-centric affordance knowledge: \oak base, and the other is the human-centric interaction knowledge: \ink base.
Once we decide the composition of \oakink, three questions shall be answered.
\begin{enumerate*}[label={\arabic*)},font={\bfseries}]
  \item How to represent the objects' affordances?
  \item How to record human experiences on manipulating the objects based on the affordances?
  \item How to transfer the recorded interactions to those objects with similar affordance?
\end{enumerate*}
To address these questions,
we describe the construction of the \oak base in \cref{sec:oakn}, present how we record and annotate the human demonstration in \cref{sec:ikn},
and introduce a novel interaction knowledge transfer method in \cref{sec:transfer}.
Finally, we provide the statistics and analysis in \cref{sec:analy}.

\vspace{-1mm}\subsection{Object-Centric Affordance Knowledge Base}\label{sec:oakn}
\begin{figure}[t]
  \begin{center}
    \includegraphics[width=1.0\linewidth]{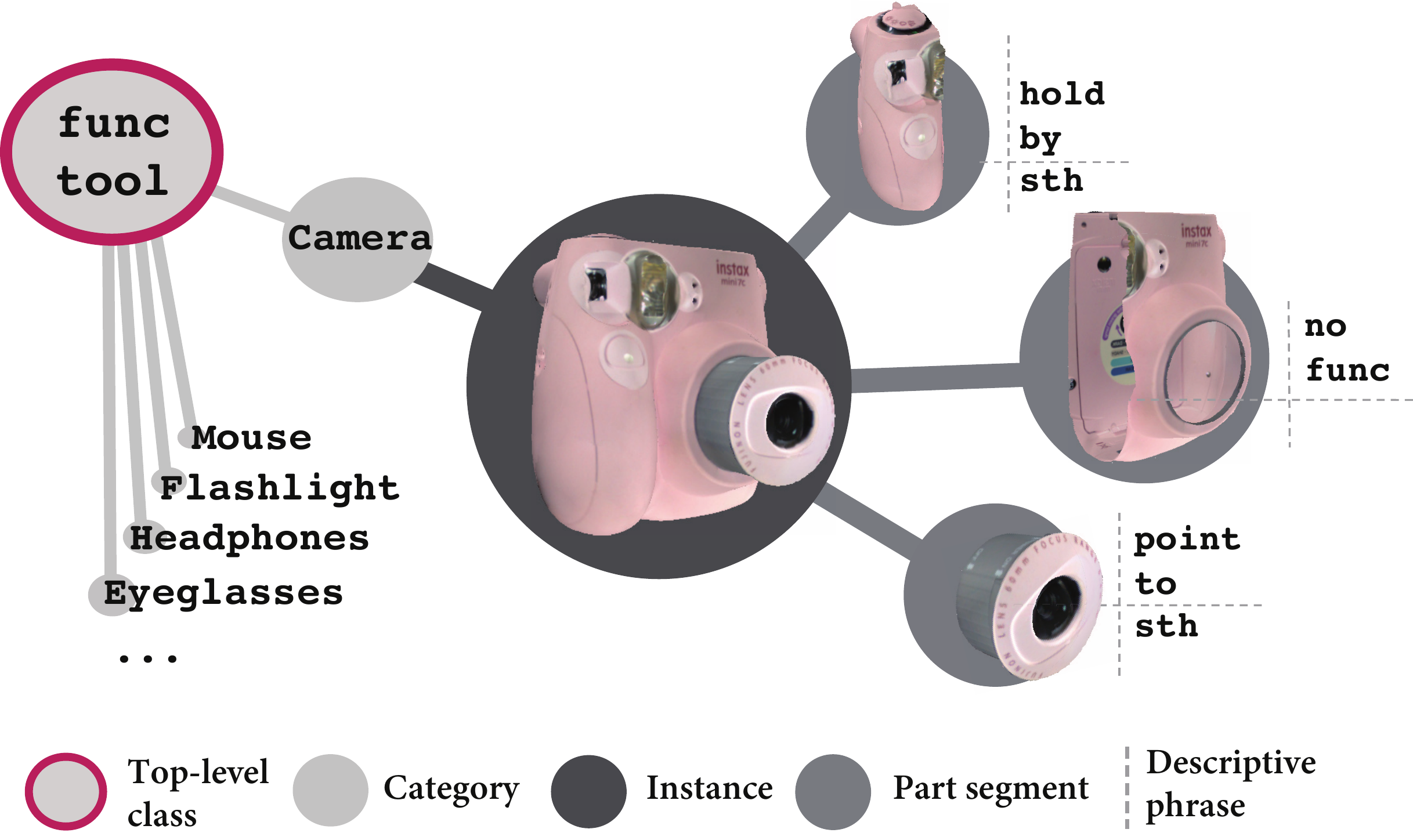}
  \end{center}
  \vspace{-6mm}
  \caption{Object Affordance Knowledge Graph.}\vspace{-3mm}
  \label{fig:obj_affod}
\end{figure}

\vspace{-1.5mm}We focus on objects that are commonly appeared in our daily life and are designed for single-hand manipulations.
We collect a total of 1,800 household objects for these purposes.
The source of these objects are four-fold:
\begin{enumerate*}[label={\arabic*)}]
  \item self-collected from online vendors,
  \item ShapeNet models,
  \item YCB objects, and
  \item ContactDB objects,
\end{enumerate*}
in which we observe  diverse object categories, shapes, and affordances.
We organize all the objects into a knowledge graph (\cref{fig:obj_affod}).
The knowledge graph, as well as the objects' scanned models form the main body of \oak base.
Next, we will elaborate on \textit{how we arrange the objects} (taxonomy) and \textit{how we describe the affordance of the objects} (attribute).
The taxonomy and attribute in \oak base should achieve \textbf{1) consensus} that the consistent classification shall be made by a group of people based on their common experience,  and have \textbf{2) scalability} that new objects and new attributes can be easily extended to the current knowledge base.
After exhaustively reviewing the objects in the above datasets, we found the taxonomy and the description of attributes can be concluded within limited categories.

\vspace{1mm}\noindent\textbf{Taxonomy.}
We adopt a taxonomy that groups \oak base objects into two levels of classifications.
We define the top-level classification that consists of two classes, namely  manipulation tools ({\tt \textbf{maniptool}}) and function tools ({\tt \textbf{functool}}). Their definitions are as follows:
\vspace{-1.5 mm}\begin{itemize}[font={\bfseries}, leftmargin=*]
  \setlength\itemsep{-1 mm}
  \item {\tt \textbf{maniptool}} class contains tools that are used to manipulate (affect) other entities. These objects usually contain a handle (for grasping) and an end effector (for affecting other entities). (\eg mug, knife, pincer  and drill);
  \item {\tt \textbf{functool}} class manages tools that usually have a self-contained function and do not necessarily require end effectors. (\eg camera and headphone);
\end{itemize}\vspace{-1mm}
Within the top-level classifications, we arrange the objects based on the WordNet \cite{miller1995wordnet} categories as the sub-level classification. The total number of categories in the \oak base is 32. We list all the categories in \supp.

\vspace{1mm}\noindent\textbf{Attribute.}
The notion of ``affordance'' was introduced by Gibson \cite{gibson2014ecological} as the characteristics of the functional properties of objects.
Later in CV and robotics community, the affordance has been used with different formulations, such as graspable area \cite{kokic2017affordance, kokic2020learning}, grasp types\cite{corona2020ganhand}, part segmentation \cite{mo2021where2act, AffordanceNet18}, contact region \cite{brahmbhatt2019contactgrasp}, and action-effects \cite{fang2020learning}. In this paper, we denote the ``affordance'' as the functionality of object.
The affordance is represented by a set of \textit{attribute}s. Each \textit{attribute} contains a part segmentation with one or several descriptive phrase(s) that describe the part's functionality.
For example, given a knife with two parts: blade and handle,
we assign the phrase \textit{$\langle$cut, something$\rangle$} to the blade, and the phrase \textit{$\langle$handled (by), something$\rangle$} to the handle.
We invite 10 volunteers with different backgrounds and ask them to firstly make a phrase of: $\langle$\textit{verb} (+ \textit{prep}), \textit{something}$\rangle$ to describe each part of the objects. We only focus on parts with functionality. Hence for parts that may not have function, we ask volunteers to judge and assign them as $\langle$\textit{no~function}$\rangle$.
We also encourage volunteers to conclude the part-level similarity across different object categories.
In the beginning, we create an empty candidate phrase pool.
When a new phrase was initially proposed on a certain part,
we first check whether it has a duplicated meaning in the candidate pool.
Then, we seek the consensus among all the volunteers on whether to replace or add it.
Finally, we gather all the phrases, conclude their meaning and vote for their occurrence on each part.
Through exhaustively reviewing all the 32 object categories, we conclude total 30 phrases as the final attribute phrases.
We list all the attribute phrases in \supp.

\subsection{Human-Centric Interaction Knowledge Base} \label{sec:ikn}
\vspace{-1.5mm}In this section, we elaborate on how we collect human demonstration and construct the \ink base.
We first introduce the hardware setup for efficient recording in \cref{sec:hardware}, provide a protocol for data acquisition in \cref{sec:data_acqui} and describe the details of data annotation in \cref{sec:data_anno}.

\vspace{-3 mm}\subsubsection{Hardware Setup}\label{sec:hardware}
\begin{figure}[t]
  \begin{center}
    \includegraphics[width=0.95\linewidth]{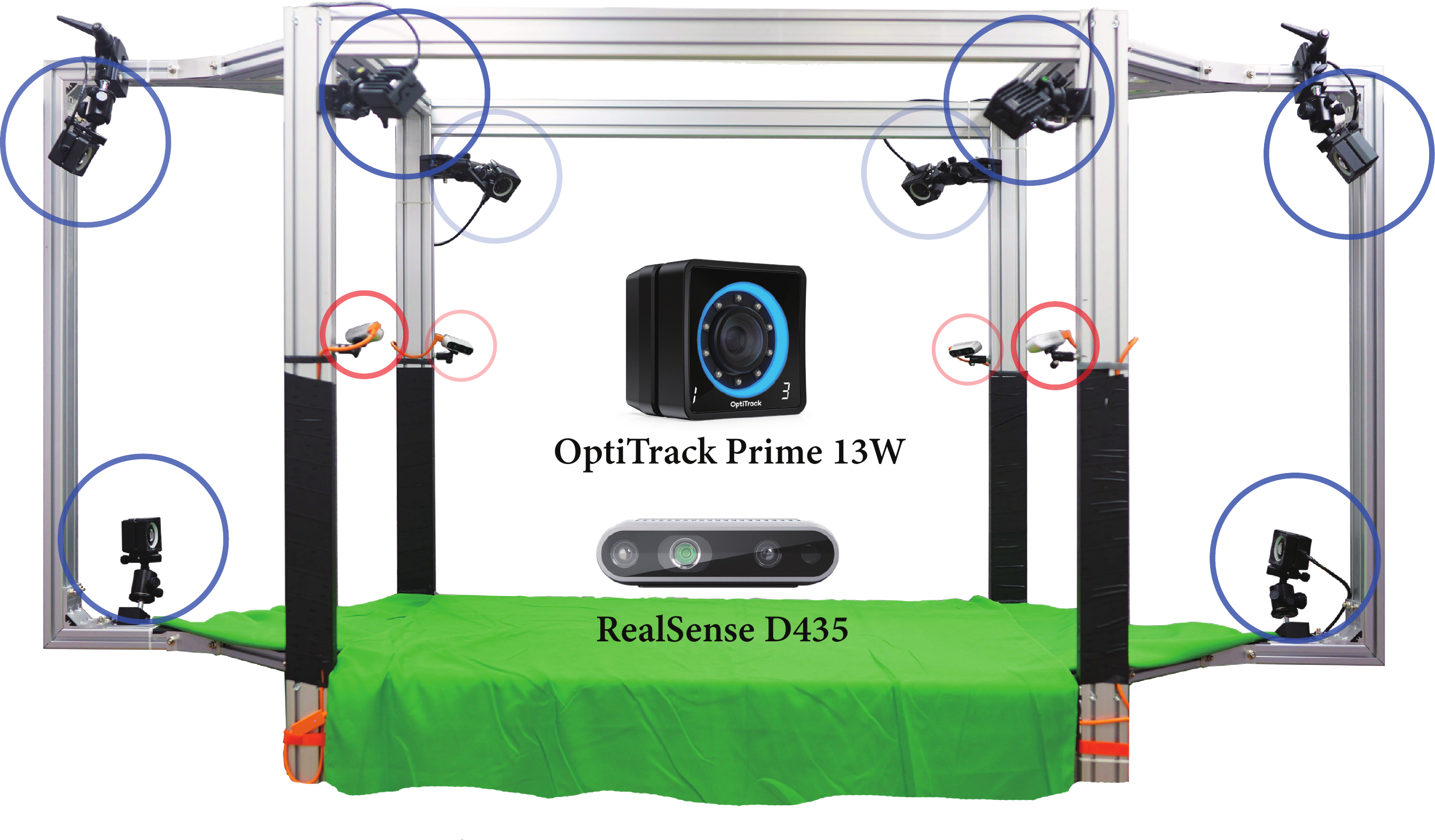}
  \end{center}\vspace{-6mm}
  \caption{Our data collection platform with 4 RGB-D cameras (red circle) and 8 infrared MoCap cameras (blue circle).}\vspace{-4 mm}
  \label{fig:hardware}
\end{figure}

\vspace{-2 mm}The data collection platform consists of a multi-camera system (MulCam) and an infrared motion capture system (MoCap). The MulCam system consists of 4 RealSense D435 cameras that are used to record the image-based interaction sequence. The MoCap system consists of 8 Optitrack Prime 13W cameras that are used to track the object's motion during the interaction. We synchronized all the cameras in both sensor systems and calibrated the transformation between the MulCam system: \camsys and MoCap system: \capsys.
Our platform is shown in \cref{fig:hardware}.
All the sensors are rigidly mounted on the edges of a $1.5 \times 1.2 \times 1$ m$^3$ cuboidal area, which enables the subject to freely interact with objects or other subjects without interference.

\vspace{-3 mm}\subsubsection{Interaction Sequence Acquisition}\label{sec:data_acqui}
\vspace{-2 mm}We invited 12 subjects and recorded their interactions with the given objects.
Each subject is assigned a subset from the object database.
A director will firstly elaborate on the \textit{attributes} of each object and confirm the acknowledgment of these \textit{attributes} among all the subjects.
Then the subjects are asked to start from a hand pose lying flat on the table, pick up the assigned object, and finish the action with a given intent.
For each object, we collect up to 5 intents, namely \textit{use}, \textit{hold}, \textit{lift-up}, \textit{hand-out}, and \textit{receive}. The intent: \textit{use} requires the subject to perform an action that makes use of the object's \textit{attribute(s)}.
The \textit{hold} requires the subject to perform a steady grasp of the object.
The \textit{lift-up} asks the subject to pick up an overturned object and place it upright.
When a subject was asked to \textit{hand-out} an object, this subject (the giver) was also paired with another subject (the receiver) to perform \textit{receive}.
The paired sequences of \textit{hand-out} and \textit{receive} constitute an action of human-to-human handover.
During handover, the giver was asked to determine where the receiver would \textit{receive} the object to \textit{use}. Meanwhile, the receiver was asked to determine how to \textit{receive} the object from the giver without mutual contact.
After each action finishes, a director will place the object with a random pose for the next action.
We record each action for 5 seconds and manually discard the idle frames.

\vspace{-4 mm}\subsubsection{Data Annotation}\label{sec:data_anno}
\vspace{-2 mm}During the entire course of the human demonstration,
we are particularly interested in the \textit{poses} and \textit{contact pattern} of hand and object,
as they embrace the human experience of manipulating objects.

\begin{figure}[t]
  \begin{center}
    \includegraphics[width=0.95\linewidth]{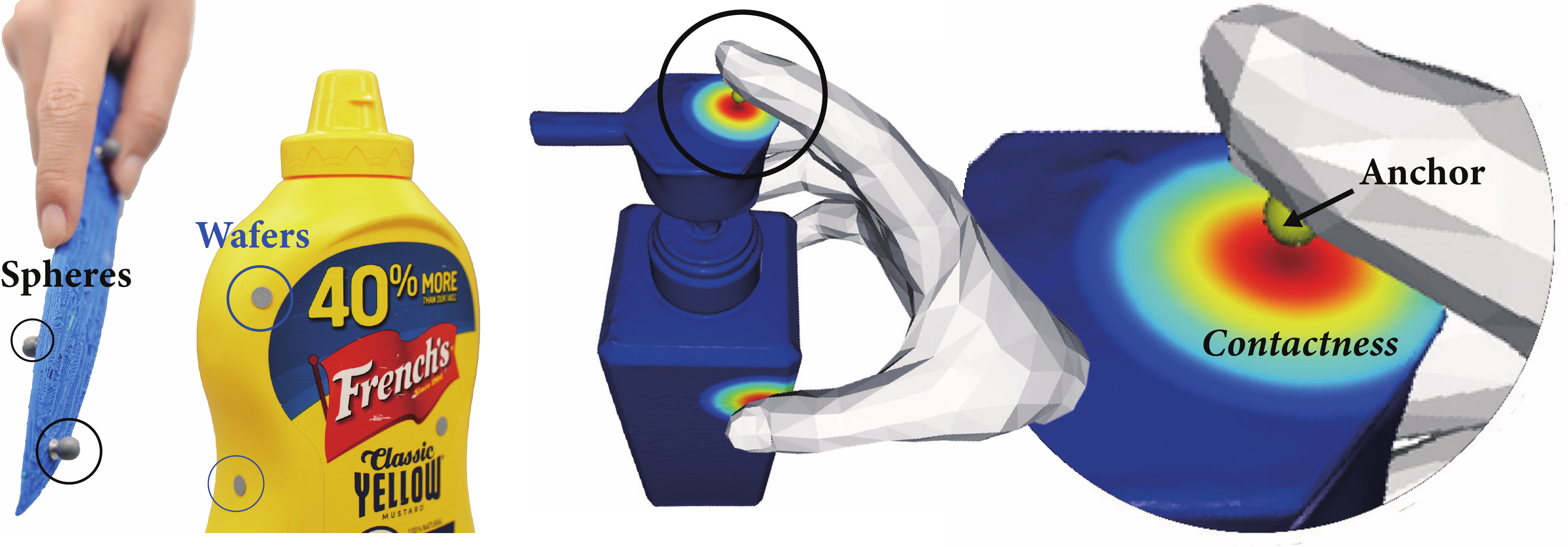}
  \end{center}\vspace{-6mm}
  \caption{Illustration of the reflective markers (left two) for tracking object motion and the \textit{contactness} that describes physical contact region (right two).}\vspace{-7mm}
  \label{fig:data_anno}
\end{figure}

\vspace{1mm}\noindent\textbf{Object Pose.}
We track the object's 6 DoF pose by tracking the surface-attached reflective markers (\cref{fig:data_anno} left) in the MoCap system \capsys.
Then,
we transform the object pose from \capsys
to the MulCam system \camsys by the system calibration.

\vspace{1mm}\noindent\textbf{Hand Pose and Geometry.}
We rely on manually labeled 2D hand keypoints from multi-views to acquire the 3D hand joints annotation.
Following the practice in \cite{chao2021dexycb}, we set up an annotation task on an online crowd-sourcing platform and require workers to locate every keypoint in all 4 views of all the assigned frames.
We adopt the standard 21 hand keypoints following the orders and locations defined in \cite{simon2017hand}.
To describe hand pose and geometry in 3D space, we use the MANO hand model \cite{romero2017embodied}. MANO represents an articulated and deformable hand by \textit{pose} $\bm{\theta} \in \mathbb{R}^{16\times3}$ and \textit{shape} $\bm{\beta} \in \mathbb{R}^{10}$ parameter.
Later in the paper, we denote ``hand pose'' as the 21 joint positions: $\bm{P}_h \in \mathbb{R}^{21\times3}$ in the \camsys system. With $\bm{\theta}$ and  $\bm{\beta}$, we can recover the hand pose $\bm{P}_h$ and mesh vertices $\bm{V}_h \in \mathbb{R}^{778 \times 3}$ through a differentiable MANO layer: $\mathcal{M(\cdot)}$ \cite{hasson2019obman}.
Solving the  $\bm{P}_h$ and $\bm{V}_h$ are formulated as an optimization task minimizing several hand-crafted cost functions.
In the main paper, we only describe the core term: 3D-2D keypoints re-projection error among multi-views.
For other auxiliary costs such as geometrical consistency, temporal smoothing, and silhouette constraint, please visit \supp.
Let $\bm{\hat{p}}_{j,v}$ be the $j$-th 2D hand keypoint annotation in the $v$-th view, $\bm{{P}}_{h,j}$ be the $j$-th 3D hand pose estimation, and let $\bm{T}_{v}$, $\bm{K}_v$ be the extrinsic and intrinsic of the camera of $v$-th view, we define the re-projection cost as:
\begin{equation}
  \setlength\abovedisplayskip{0pt}
  \setlength\belowdisplayskip{5pt}
  E_{\rm{repj}}=\frac{1}{\sum w_{j, v}} \sum^{N_v}_{v=1} \sum^{N_j}_{j=1} w_{j, v} \Big\| \bm{K}_v \bm{T}_v \bm{{P}}_{h,j} - \bm{\hat{p}}_{j,v} \Big\|^2_2,
  \label{eq:repj_cost}
\end{equation}
where $w_{j, v}$ indicates the visibility of the keypoint $\bm{\hat{p}}_{j,v}$.

\vspace{1mm}\noindent\textbf{Contact and Stress Pattern.}\label{par:contact}
Given the accurate hand and object poses, we can derive the per-hand-part contact region on object surfaces.
We adopt the 17 hand parts segmentation and part-level anchor location in Yang \etal \cite{yang2021cpf}.
Based on the efficient contact heuristic in GRAB \cite{taheri2020grab}, we automatically assign a part label to those vertices on the object's surface if an anchor is close to them (within the threshold of $25$ mm). The vertices with a labeled hand part form the contact regions of the object.
Physical contact commonly results in an elastic deformation of both hand and object \cite{grady2021contactopt}, in which the stress and strain will spread across the deformation area. Though MANO and rigid object model cannot reflect this behavior, we can imitate the stress pattern by adding a ring-shape spreading and decreasing value in the contact region, as we call it \textit{contactness}.
As shown in \cref{fig:data_anno} right,
the \textit{contactness} takes the maximum value $1$ at the point closest to a certain anchor, centrally decreasing as the distance increases, and finally becomes $0$ when the distance is greater than $25$ mm.
We delay the demonstration on the use of \textit{contactness} until \cref{sec:pose_refine}.

\begin{figure*}[t]
  \begin{center}
    \includegraphics[width=0.92\linewidth]{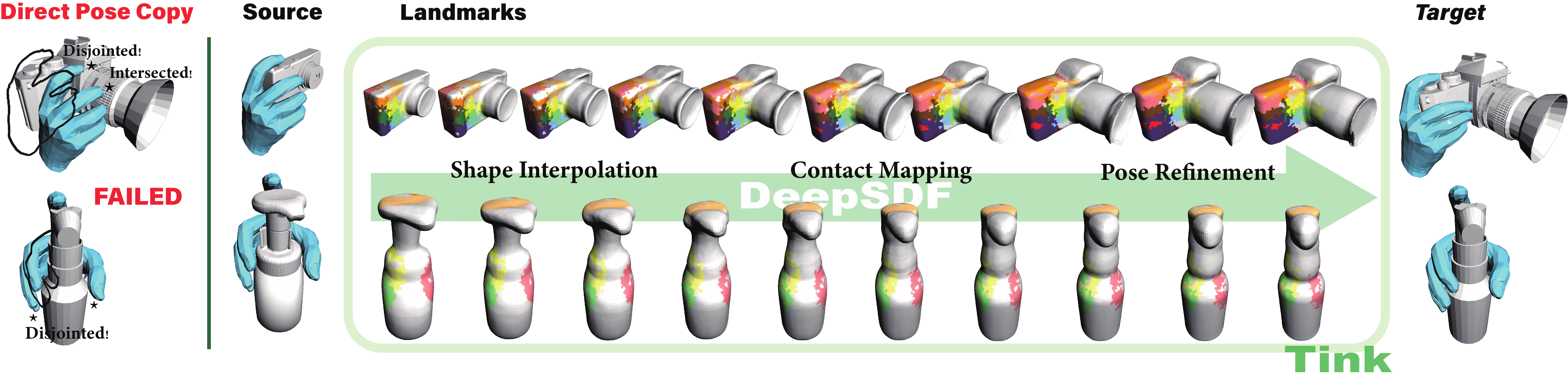}
  \end{center}\vspace{-7.0mm}
  \caption{(\textit{Best view in color}) Illustration of our \tink pipeline. ``Direct pose copy'': copying hand pose ($\bm{\theta}$) in \textit{source} object system to \textit{target} object without refinement. This pose copying usually suffers from unnatural disjointedness or intersection due to shape variant.}\vspace{-4.0mm}
  \label{fig:tink}
\end{figure*}

\vspace{-1 mm}\subsection{\textbf{\tink}: Transferring Interaction Knowledge} \label{sec:transfer}
\vspace{-1.5 mm} 
This section describes how we transfer the hand's interactions with the real-world objects (recorded in human demonstration) to the virtual counterpart objects  (collected in \oak base) of the same category.
The transferred interactions should be consistent with those collected regarding contact, pose, intent, and human perception.
However, as different objects vary in shapes and sizes, direct pose copying (as shown in \cref{fig:tink} left) would fail in most cases.
To address this issue, we propose a hybrid learning-fitting method: \textbf{\tink} for \textbf{T}ransferring the \textbf{in}teraction \textbf{k}nowledge.
\tink consists of three sequential modules, namely shape interpolation, contact mapping, and pose refinement.

We refer to the objects that have been recorded in real-world human demonstrations as the \textit{source} objects, and the virtual objects in \oak as the \textit{target} objects.
As a recorded sequence only has one type of hand-object interaction (handover sequence has two), we manually select 1 (or 2) steady interacting pose(s) to represent each sequence. 
These selected hand poses are the \textit{source} poses for interaction transfer. 
Later, we call the set of those selected interactions \oakink-Core.

\vspace{-4 mm}\subsubsection{Implicit Shape Interpolation}\label{sec:shape_interp}
\vspace{-2 mm}
Once we decide to transfer the interaction from one \textit{source} object to another \textit{target} object,
an instant question is how to express the object shape and perform continuous shape deformation.
To answer this, we first represent the object shape as an implicit function (SDF, signed distance function), as SDF is naturally continuous.
Now the question is how to perform the shape interpolation between the SDF of \textit{source} and \textit{target}.
To address this, we adopt a neural generative model: DeepSDF \cite{park2019deepsdf} that maps complex 3D shapes into a continuous latent space.
Using DeepSDF has three advantages.
1) We can acquire a compact representation of complex shape, namely the shape vector;
2) We can perform accurate shape interpolation by interpolating the shape vectors in the latent space;
3) Later in the \cref{sec:pose_refine}, we can mitigate the hand-object interpenetration by penalizing the negative query positions (\cref{eq:intep_cost});

We firstly train a DeepSDF model on all the \textit{source} and \textit{target} object SDFs of a certain category.
Then for the $i$-th \textit{source} object SDF: $\mathbfcal{O}^s_i$ and $j$-th \textit{target} object SDF: $\mathbfcal{O}^t_j$, we perform linear interpolation between their latent shape vector: $\bm{o}^s_i$ and $\bm{o}^t_j$.
During the interpolation, we sample $N_{itpl}$ equally spaced quantiles as landmarks.
Finally, We decode the shape vector at landmarks to its SDF and reconstruct a mesh model by Marching Cubes \cite{lorensen1987marching}.
The $N_{itpl}$ artificial objects constitute a path connecting the \textit{source} and \textit{target}.

\vspace{-4 mm}\subsubsection{Explicit Contact Mapping}
\vspace{-2 mm}As shown in \cref{fig:tink} left, directly copying the hand pose would fail.
We need to find a piece of consistent information shared among the \textit{source}, the \textit{target}, and the $N_{itpl}$ landmarks along the path.
Compared to pose, contact regions are more invariant against shape deformation.
We start mapping contact regions from the \textit{source} object, sequentially pass through the $N_{itpl}$ landmarks, and finally reach the \textit{target} object.
As long as the interval between each two landmarks is small enough, we can neglect shape variation between the $i$-th and ($i$+$1$)-th objects.
The contact mapping is illustrated in \cref{fig:tink} (contact regions of different finger parts are painted with different colors).
Considering the trade-off between efficiency and accuracy, we empirically find that $N_{itpl} = 10$ is sufficient enough.
We map the contact label of a vertex on the $i$-th object to its closest vertex on the ($i$+$1$)-th object.
At each $i$ $(0\le i < N_{itpl})$ step, we adopt the iterative closest point (ICP) to link the corresponding vertices.

\vspace{-4 mm}\subsubsection{Iterative Pose Refinement}\label{sec:pose_refine}
\vspace{-2 mm}In the last module, we map the interacting hand pose of the \textit{source} object to its counterpart \textit{target} objects.
As we express the knowledge of interaction as the semantics on the object surface, namely the contact regions (recall \cref{par:contact}), pose mapping is conducted by enforcing the contact consistency between the \textit{source} and \textit{target} object.
We formulate pose mapping as an iterative optimization. The variables during optimization are the pose $\bm{\theta}$, shape $\bm{\beta}$, and wrist position $\bm{P}_{h_0}$ of a newly transferred hand.

We start to attract the anchors on the hand surface to its corresponding contact regions on the \textit{target} object.
Let the anchors of total 17 hand regions be $\mathbfcal{A} = \{\bm{A}_i\}_{i=1}^{17}$,
the vertices on the object surface that corresponds to the anchor $\bm{A}_i$ be
 $\mathbfcal{V}_h^{(i)} = \{ \bm{V}^{(i)}_{h,j} \}$,
and the \textit{contactness} between  $\bm{A}_i$ and $\bm{V}_{h,j}^{(i)}$ be $\gamma_{ij}$.
The contact consistency cost is expressed as:
\vspace{-1mm}\begin{equation}
  \small
  \setlength\abovedisplayskip{5pt} 
  \setlength\belowdisplayskip{5pt}
  E_{\rm{consis}} = \frac{1}{\sum{\gamma_{ij}}}\sum_{\bm{A}_i} \sum_{\bm{V}_{h,j}^{(i)}} \gamma_{ij} \big \| \bm{A}_i -  \bm{V}_{h,j}^{(i)} \big \|^2_2,
  \label{eq:consist}\vspace{-2mm}
\end{equation}

Direct optimization on the joints' rotations $\bm{\theta}$ is prone to anatomical abnormality. Hence we adopt the axial adaptations from Yang \etal \cite{yang2021cpf} and constrain the rotation axes and angles.
Let $\bm{a}_j$ and $\phi_j$ be the axial and angular components of the $j$-th joint's rotation, the $\mathbf{n}_j^{t}$ and $\mathbf{n}_j^{s}$ be the pre-defined \textit{twist} and \textit{splay} direction. The anatomical cost is defined as:
\begin{equation}
  \small
  \setlength\abovedisplayskip{5pt}
  \setlength\belowdisplayskip{5pt}
  E_{\rm{anat}}=\sum_{j \in \text{all}} \Big(\bm{a}_j \cdot \mathbf{n}_j^{t} + \max\big( (\phi_j - \frac{\pi}{2}), 0\big)\Big)+ \sum_{j \notin \text{MCP}} \bm{a}_j \cdot \mathbf{n}_j^{s},
  \label{eq:anat_cost}\vspace{-1mm}
\end{equation}
where ``MCP'' indicates the five Metacarpal joints.

In order to control the hand-object interpenetration, we also introduce an interpenetration cost to penalize those hand vertices inside the object surface:
\begin{equation}
  \small
  \setlength\abovedisplayskip{5pt}
  \setlength\belowdisplayskip{5pt}
  E_{\rm{intp}}=\sum_{\bm{V}_{h, j}} - \min \Big(\text{SDF}_{\bm{\mathbfcal{O}}}(\bm{V}_{h, j}) , 0\Big),
  \label{eq:intep_cost}\vspace{-1mm}
\end{equation}
where the $\text{SDF}_{\mathbfcal{O}} (\cdot)$ calculates the signed distance value of a 3D hand vertex $\bm{V}_{h, j}$ to an object's SDF: $\mathbfcal{O}$ provided at shape interpolation (\cref{sec:shape_interp}).
The total optimization problem is:
\begin{equation}
  \small
  \setlength\abovedisplayskip{5pt}
  \setlength\belowdisplayskip{8pt}
  \bm{V}_h, \bm{P}_h \longleftarrow \mathop{\rm{argmin}}_{\bm{\theta}, \bm{\beta}, \bm{P}_{h_0}} \big(E_{\rm{consis}} + E_{\rm{anat}} + E_{\rm{intp}}\big),
  \label{eq:hand_annot}\vspace{-2mm}
\end{equation}
where  $\bm{V}_h, \bm{P}_h = \mathcal{M}(\bm{\theta}, \bm{\beta}) + \bm{P}_{h_0}$. We run 1,000 iterations per \textit{source-target} pair. The whole pipeline is implemented in PyTorch with Adam solver.

\noindent\textbf{Perceptual Evaluation.} Finally, all the transferred interactions are sent to 5 volunteers for perceptual evaluation.
Given the \textit{source} object and its interacting hand pose as a reference, the volunteers are asked to make a judgment on whether the transferred hand pose on \textit{target} object demonstrates the same intents and satisfies visual plausibility.
We only select the interactions that achieve consensus on plausibility among the 5 volunteers.

\begin{table*}[ht]
    \centering
    \footnotesize
    \setlength{\tabcolsep}{1.30pt}
    \resizebox{0.9\textwidth}{!}{
      \begin{tabular}{l|cccccccc|ccccccc}
        \toprule
        \multirow{2}{*}{Dataset}                     & \multirow{2}{*}{\minitab[c]{mod.}} & \multirow{2}{*}{reslution} & \multirow{2}{*}{\#frame} & \multirow{2}{*}{\#subj} & \multirow{2}{*}{\#obj} & \multirow{2}{*}{\#views} & \multirow{2}{*}{\#inten} & \multirow{2}{*}{\#intact} & \multirow{2}{*}{\minitab[c]{real /                                                                                  \\syn.}} &  \multirow{2}{*}{\minitab[c]{label\\method}} & \multirow{2}{*}{\minitab[c]{intac.\\inten}} & \multirow{2}{*}{\minitab[c]{obj\\pose}} & \multirow{2}{*}{\minitab[c]{dynamic\\ intac.}}&  \multirow{2}{*}{\minitab[c]{hand-\\over}} &  \multirow{2}{*}{\minitab[c]{hand-obj\\contact.}} \\
                                                     &                                    &                            &                          &                         &                        &                          &                          &                           &                                    &          &             &             &             &             &             \\
        \midrule
        ObMan~\cite{hasson2019obman}                 & RGBD                               & $256 \times 256$           & 154K                     & 20                      & 3K                     & 1                        & --                       & --                        & syn                                & simulate & \redcross   & \greencheck & \redcross   & \redcross   & \redcross   \\
        YCBAfford~\cite{corona2020ganhand}           & RGB                                & --                         & 133K                     & 1                       & 21                     & 1                        & --                       & 367                       & syn                                & simulate & \redcross   & \redcross   & \redcross   & \redcross   & \redcross   \\
        \midrule
        FPHAB~\cite{FirstPersonAction_CVPR2018}      & RGBD                               & $1920\times 1080$          & 105K                     & 6                       & 4                      & 1                        & 3                        & 273                       & real                               & marker   & \greencheck & \greencheck & \greencheck & \redcross   & \redcross   \\
        HO3D~\cite{hampali2020ho3dv2}                & RGBD                               & $640 \times 480$           & 78K                      & 10                      & 10                     & 1-5                      & --                       & 68                        & real                               & auto     & \redcross   & \greencheck & \greencheck & \redcross   & \redcross   \\
        ContactPose~\cite{brahmbhatt2020contactpose} & RGBD                               & $960 \times 540$           & 2991K                    & 50                      & 25                     & 3                        & 2                        & 2.3K                      & real                               & auto     & \greencheck & \greencheck & \redcross   & \redcross   & \greencheck \\
        GRAB~\cite{taheri2020grab}                   & Mesh                               & --                         & 1624K                    & 10                      & 51                     & --                       & 4                        & 1.3K                      & real                               & marker   & \greencheck & \greencheck & \greencheck & \redcross   & \greencheck \\
        DexYCB~\cite{chao2021dexycb}                 & RGBD                               & $640 \times 480$           & 582K                     & 10                      & 20                     & 8                        & --                       & 1K                        & real                               & crowd    & \redcross   & \greencheck & \greencheck & \redcross   & \redcross   \\
        H2O~\cite{kwon2021h2o}                       & RGBD                               & $1280\times 720$           & 571K                     & 4                       & 8                      & 5                        & 7                        & 1.8K                      & real                               & auto     & \greencheck & \greencheck & \greencheck & \redcross   & \redcross   \\
        \midrule
        \textbf{Ours \oakink-Image}                  & RGBD                               & $848 \times 480$           & 230K                     & 12                      & \textbf{100}           & 4                        & 5                        & 1K                        & real                               & crowd    & \greencheck & \greencheck & \greencheck & \greencheck & \greencheck \\
        \textbf{Ours \oakink-Shape}                  & Mesh                               & --                         & --                       & 12                      & \textbf{1,700}         & --                       & 5                        & \textbf{49K}              & real                               & \tink    & \greencheck & \greencheck & \redcross   & \greencheck & \greencheck \\
  
        \toprule
      \end{tabular}
    }\vspace{-3mm}
    \caption{\small Comparison of our \oakink with the publicly available datasets of hand-object interactions.}
    \vspace{-3mm}
    \label{tab:dataset_comp}
  \end{table*}

\vspace{-1 mm}\subsection{Dataset Analysis} \label{sec:analy}
\vspace{-2 mm}In this section, we provide the statistic and analysis of \oakink.
As a summary, we collected 230K image frames of 12 subjects performing up to 5 intent-oriented interactions with total 100 real-world objects of 32 categories, and transferred the interactions to the rest of 1,700 virtual counterpart objects. The total number of distinct hand-object interactions is 50,000.
We denote the image-based dataset as \textbf{\oakink-Image}. 
We select 1 (or 2) representative hand-object interaction for each image sequence and denote their collection as \oakink-Core. 
All the selected and transferred interactions constitute another geometry-based dataset: \textbf{\oakink-Shape}.
We make a comprehensive comparison with the existing hand-object datasets in \cref{tab:dataset_comp}, and visualize hand pose and contact distribution in \supp. 

\vspace{0.5mm}\noindent\textbf{Image Dataset Cross Validation.}
To evaluate the merit of \oakink-Image, we perform cross-dataset validation in \cref{tab:cross_dataset_validation}. We train an image-based 3D pose estimation model \cite{sun2018integral} separately on three training sets: HO3D, \oakink-Image, and their mixture, and report the hands' MPJPE on DexYCB testing set. 
We observe consistent MPJPE improvements on the model trained on \oakink-Image (alone and mixture), verifying that \oakink-Image complements HO3D dataset and improves the network model. 
\begin{table}[h]
    \renewcommand{\arraystretch}{1.0}
    \centering
    \scriptsize
    \resizebox{0.68\linewidth}{!}{
        \setlength{\tabcolsep}{4.3pt}{
            \makeatletter\def\@captype{table}\makeatother
            \setlength{\tabcolsep}{1mm}{
                \begin{tabular}{l|c|c}
                    \toprule
                    \textbf{Train}         & \textbf{Test} & {\footnotesize \textbf{MPJPE} (\textit{mm}) $\downarrow$ } \\ 
                    \midrule
                    1) HO3D                   & DexYCB        & 55.38                                       \\ 

                    \cellcolor{Gray} 2) \oakink-Image        & \cellcolor{Gray} DexYCB        & \cellcolor{Gray}44.81                                       \\ 

                    \cellcolor{DGray} \textbf{1) \& 2) mixture} & \cellcolor{DGray} DexYCB        & \cellcolor{DGray}\textbf{39.70}                                       \\ 
                    \bottomrule
                \end{tabular}
            }
        }}
    \vspace{-2 mm}
    \caption{Cross dataset validation on \oakink-Image}
    \vspace{-1 mm}
    \label{tab:cross_dataset_validation}
\end{table}

\vspace{0.5mm}\noindent\textbf{Geometry Dataset Qualities.}
To evaluate the quality of \oakink-Shape, we inspect several physical-based metrics that assess the feasibility and stability of grasps. 
We also compare those metrics with the other three datasets: FPHAB \cite{FirstPersonAction_CVPR2018}, GRAB (GrabNet split) \cite{taheri2020grab}, and HO3D \cite{hampali2020ho3dv2}, representing three different data annotation methods: active magnetic transmitter, passive reflective markers, and automatic marker-less, respectively. \cref{tab:dataset_quality} shows that \oakink-Shape exhibits high physical-based qualities. 

\vspace{-2mm}\begin{table}[h]
    \centering
    \scriptsize
    \resizebox{0.95\linewidth}{!}{
        \setlength{\tabcolsep}{4.3pt}{
            \makeatletter\def\@captype{table}\makeatother
            \setlength{\tabcolsep}{1mm}{
                \begin{tabular}{l|c|c|c|c|c}
                    \toprule
                    \multirow{1}{*}{\textbf{Metrics} }             & \multirow{1}{*}{\textbf{$\star$-Core}} & \multirow{1}{*}{\textbf{$\star$-Shape}} & \multirow{1}{*}{FPHAB} & \multirow{1}{*}{GRAB} & \multirow{1}{*}{HO3D} \\
                    \midrule
                    \textit{Penet. Depth. cm}$\downarrow$          & 0.18                                   & \textbf{0.11}                           & 1.95                   & 2.53                  & 1.16                  \\
                    \textit{Solid Intsec. Vol. cm$^3$}$\downarrow$ & 1.03                                   & \textbf{0.62}                           & 22.87                  & 7.61                  & 2.08                  \\
                    \textit{Sim. Disp. Mean cm}$\downarrow$        & 0.98                                   & \textbf{0.94}                           & 6.60                   & 2.04                  & 1.91                  \\
                    \textit{Sim. Disp. Std cm}$\downarrow$         & 1.74                                   & \textbf{1.62}                           & 5.34                   & 3.17                  & 2.88                  \\
                    \bottomrule
                \end{tabular}
            }
        }}
    \vspace{-2 mm}
    \caption{Quality assessment of $\bm{\star}$(:\oakink)-Shape. To note: evaluation on PFHAB and HO3D are only conducted on frames of hand grasping objects (minimal distance $\le$ 5 mm).} \vspace{-4.0mm}
    \label{tab:dataset_quality}
\end{table}

\vspace{-4 mm}\section{Tasks and Benchmark Results}
\vspace{-2 mm}We benchmark three existing tasks (\cref{sec:hmr}-\ref{sec:graspgen}) and propose two novel tasks (\cref{sec:two_task}) on our \oakink.
The three existing tasks are:
3D hand mesh recovery (HMR, \cref{sec:hmr})\cite{lv2021handtailor, moon2020i2lmehsnet}, 3D hand-object pose estimation (HOPE, \cref{sec:hope}) \cite{hasson2020leveraging, tekin2019h+o}, and grasp pose generation (GraspGen, \cref{sec:graspgen}) \cite{taheri2020grab}. The two novel tasks are intent-based interaction generation (IntGen, \cref{sec:two_task} \textbf{A}) and  human-to-human handover generation (HoverGen \cref{sec:two_task} \textbf{B}).

\vspace{-1 mm}\subsection{Hand Mesh Recovery}\label{sec:hmr}
\begin{figure}[t]
  \begin{center}
    \includegraphics[width=1.0\linewidth]{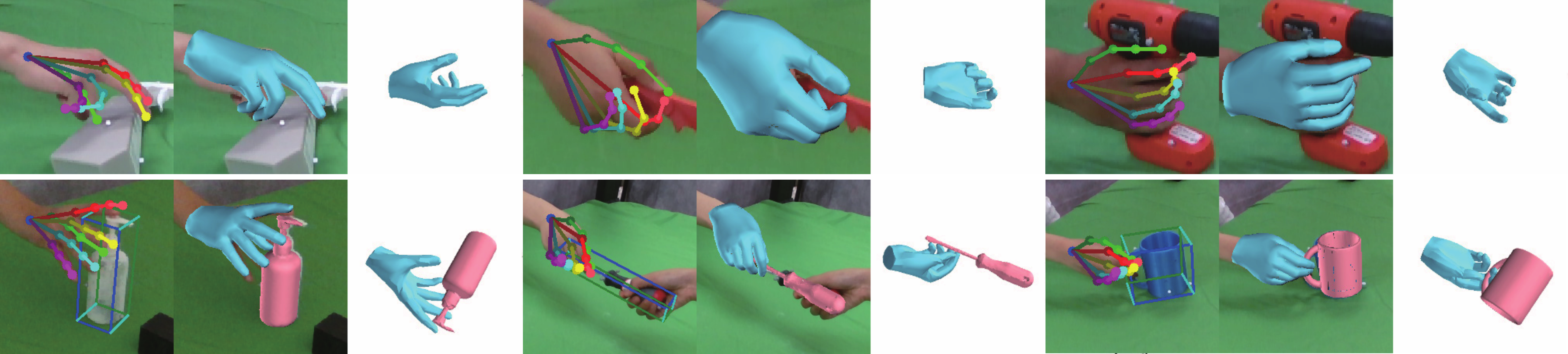}
  \end{center}\vspace{-5mm}
    \caption{Qualitative results of I2L-MeshNet\cite{moon2020i2lmehsnet} on HMR task (top row), and Hassen \etal \cite{hasson2020leveraging} on HOPE task (bottom row).}\vspace{-1mm}
  \label{fig:hope_qual}
\end{figure}

\vspace{-1 mm}The HMR task is to estimate the hand pose $\bm{P}_h \in \mathbb{R}^{21\times3}$ and geometry $\bm{V}_h \in \mathbb{R}^{778\times3}$ from a single image. 
To benchmark \oakink on this task, we first generate the train/test splits of the image frames collected in \cref{sec:ikn}. We call this image-based subset: \oakink-Image.
We randomly select one view per sequence and mark all images from this view as the test sequence, while the rest three views form the train/val sequences (train/val/test: 70\% /5\% /25\%). We call this split {\tt \textbf{SP0}} (default split).
Next, we benchmark two HMR methods: 
one is a direct image-to-vertices method: I2L-MeshNet\cite{moon2020i2lmehsnet},
and the other is a hybrid inverse kinematic method: HandTailor \cite{lv2021handtailor}.
We evaluate these methods with three metrics: mean per joint position error (\textbf{MPJPE}), percentages of correct keypoints under the curve (\textbf{AUC}) within range: [0, 50\textit{mm}], and mean per vertex position error (\textbf{MPVPE}) in wrist-relative system.
We show results on {\tt \textbf{SP0}} test set in \cref{tab:hmr} and \cref{fig:hope_qual} (top row). More quantitative results on other splits are provided in \supp.

\begin{table}[t]
    \centering
    \resizebox{0.80\linewidth}{!}{
        \footnotesize
        \setlength{\tabcolsep}{4.3pt}
        \begin{tabular}{l|l|c|c}
            \toprule
            \textbf{Splits}                   & \textbf{Methods}                & \textbf{MPJPE} (\textit{AUC}) & \textbf{MPVPE} \\
            \midrule
            \multirow{2}{*}{\tt \textbf{SP0}} & I2LMeshNet~\cite{moon2020i2lmehsnet} & 12.10  (\textit{0.784})          & 12.29             \\
                                              & HandTailor \cite{lv2021handtailor}  & 11.20  (\textit{0.884})          & 11.75             \\
            \bottomrule
        \end{tabular}
    }
    \vspace{-2.0mm}
    \caption{\small HMR results in \textit{mm}. AUC are shown in parentheses.}
    \vspace{-1mm}
    \label{tab:hmr}
\end{table}
\begin{table}[t]
    \centering
    \footnotesize
    \resizebox{0.97\linewidth}{!}{
        \setlength{\tabcolsep}{4.3pt}{
            \makeatletter\def\@captype{table}\makeatother
            \setlength{\tabcolsep}{1mm}{
                \begin{tabular}{l|c|c|cccc}
                    \toprule
                    \multirow{2}{*}{\textbf{Method}}         &
                    \multirow{2}{*}{\textbf{MPJPE}}          &
                    \multirow{2}{*}{\tabincell{c}{\textbf{MPCPE}                                                                                                                                              \\(all category)}} &
                    \multicolumn{4}{c}{${\bm{\star}}$\textbf{MPCPE} (per category)}                                                                                                                          \\
                    \cline{4-7}
                                                             &       &       & \tabincell{c}{\textit{knife}} & \tabincell{c}{\textit{lotion}} & \tabincell{c}{\textit{mug}} & \tabincell{c}{\textit{camera} } \\
                    \midrule
                    Hasson \etal \cite{hasson2020leveraging} & 27.26 & 56.09 & 68.40                         & 60.70                          & 37.26                       & 68.13                           \\
                    Tekin \etal \cite{tekin2019h+o}          & 23.52 & 52.16 & 57.29                         & 57.11                          & 35.44                       & 56.87                           \\ 
                    \bottomrule
                \end{tabular}
            }
        }}\vspace{-2mm}
    \caption{\small HOPE results in \textit{mm}. ${\bm{\star}}$: only list 4 categories.}
    \label{tab:hope}\vspace{-4mm}
\end{table}
\subsection{Hand-Object Pose Estimation}\label{sec:hope}
\vspace{-1 mm}The HOPE task is to simultaneously estimate the hand pose $\bm{P}_h$ and the object pose (rotation $\bm{R}_o \in \mathbf{SO(3)}$, center translation $\bm{t}_o \in \mathbb{R}^3 $) from a single image.
Most previous methods focused on instance-level object pose estimation. 
The object models (in the form of mesh vertices or corners) are provided as input when computing the loss during training.  
Following the same protocol,  we train and test the neural networks on the same objects.
The data split for training HOPE task follows \oakink-Image {\tt \textbf{SP0}}. 

We benchmark two representative HOPE architecture designs: Tekin \etal \cite{tekin2019h+o} and Hasson \etal \cite{hasson2020leveraging}. To note, as these two methods output the object pose in different ways, we provide adaption layers at their output. We represent object pose as the oriented 8 corners on 3D object bounding box. We evaluate these methods with two metrics: {\textbf{MPJPE}} and mean per corners position error (\textbf{MPCPE}), both in the hand wrist-relative system. We show the test set results in \cref{tab:hope} and \cref{fig:hope_qual} (bottom row).

\begin{figure}[t]
    \begin{center}
      \includegraphics[width=1.0\linewidth]{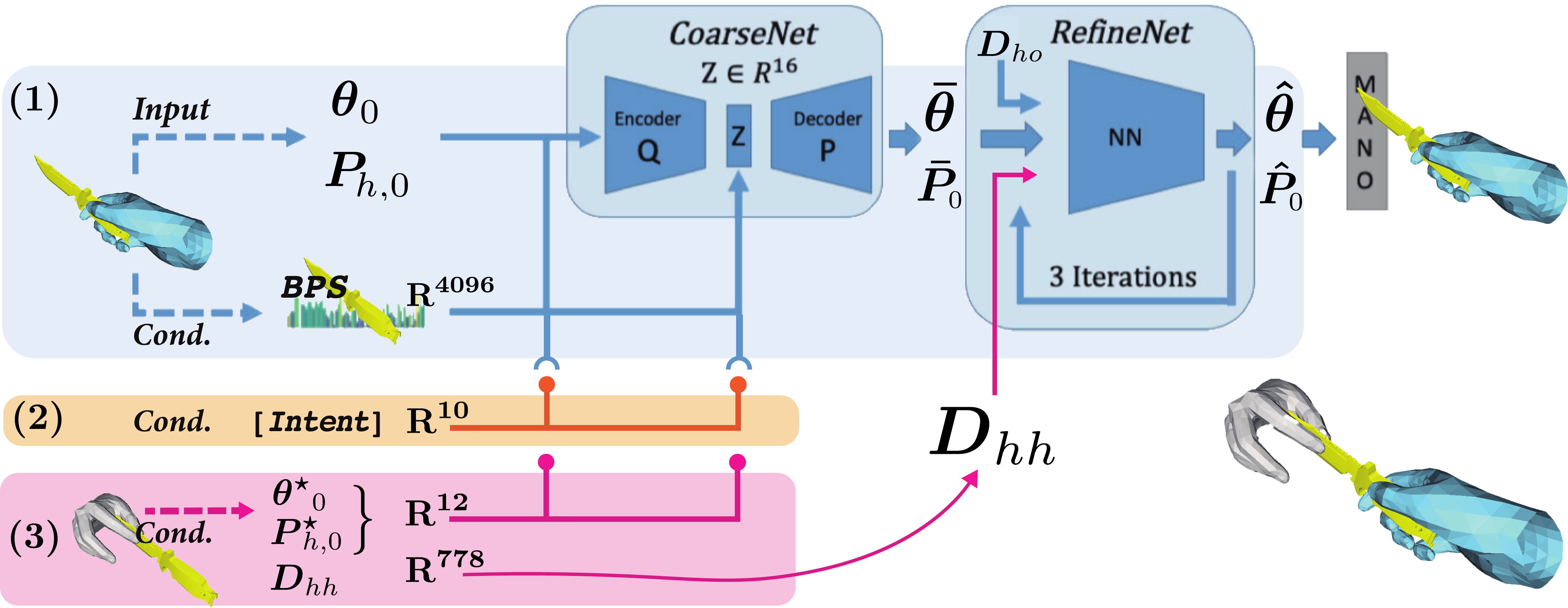}
    \end{center}\vspace{-5mm}
      \caption{\textbf{Network architectures.} (1). GrabNet; (1)+(2): Intent-based interaction generation; (1)+(3): Handover generation.}\vspace{-5mm}
    \label{fig:grabnet3}
\end{figure}

\vspace{2mm}\subsection{Grasp Pose Generation}\label{sec:graspgen}
\vspace{-2mm}The GraspGen task is to generate diverse hand poses that interact with a given object shape. 
Existing GraspGen methods \cite{karunratanakul2020grasping, taheri2020grab, jiang2021graspTTA} widely adopted a conditional VAE\cite{Sohn2015CVAE} architecture to this end. 
As shown in \cref{fig:grabnet3} (1),
the model is trained with an object shape (as BPS\cite{prokudin2019bps} $ \in \mathbb{R}^{4096}$) and its interacting hand pose ($\bm{\theta_0}$, $\bm{P}_{h_0}$) as input, and is supervised to generate the consistent hand with the input hand.
As a result, the model learns an object-conditioned hand embedding space: $\mathcal{Z}$. 
Then during the testing, given a test object, the model decodes a hand pose from its embedding space $\mathcal{Z}$. 
To benchmark our \oakink on GraspGen,
we randomly select 80\% of objects from \oak base for training, 10\% for validation, and the rest 10\% for testing. 
All the object models are paired with their interacting hand poses in group.
We denote this shape-based subset as \oakink-Shape. 

We benchmark \oakink-Shape on GrabNet \cite{taheri2020grab}, a representative method toward GraspGen. 
The evaluation metrics of GraspGen consist of four terms. 
We evaluate 1) penetration depth, 2) solid intersection volume following \cite{yang2021cpf}, and 3) simulation displacement following \cite{hasson2019obman}.
To investigate general audience's opinion about the generated poses, 
we also provide a 4) perceptual survey on the network predictions at Amazon Mechanical Turk \cite{AMT} following the practices in  \cite{karunratanakul2020grasping,taheri2020grab, jiang2021graspTTA}. We ask the workers to rate the generated hand poses with scores ranging from 1 (strongly unsatisfied) to 5 (strongly satisfied). Protocols and demonstrations of this perceptual survey are shown in \supp. We report all the four evaluation results in \cref{tab:graspgen} column 2.

\vspace{-2mm}\subsection{Two Novel Generation Tasks}\label{sec:two_task}
\vspace{-2mm}Previous GraspGen methods can only generate general grasp poses that are agnostic toward intents.
In this paper, we investigate pose generation with two applicable purposes, namely,  \textbf{A)}. to generate plausible poses with a specific intent and~  \textbf{B)}. to generate plausible poses for receiving the objects from a giver. We illustrate their network design in \cref{fig:grabnet3}. 
For implementation details, please visit \supp.

\begin{figure}[t]
    \begin{center}
      \includegraphics[width=0.9\linewidth]{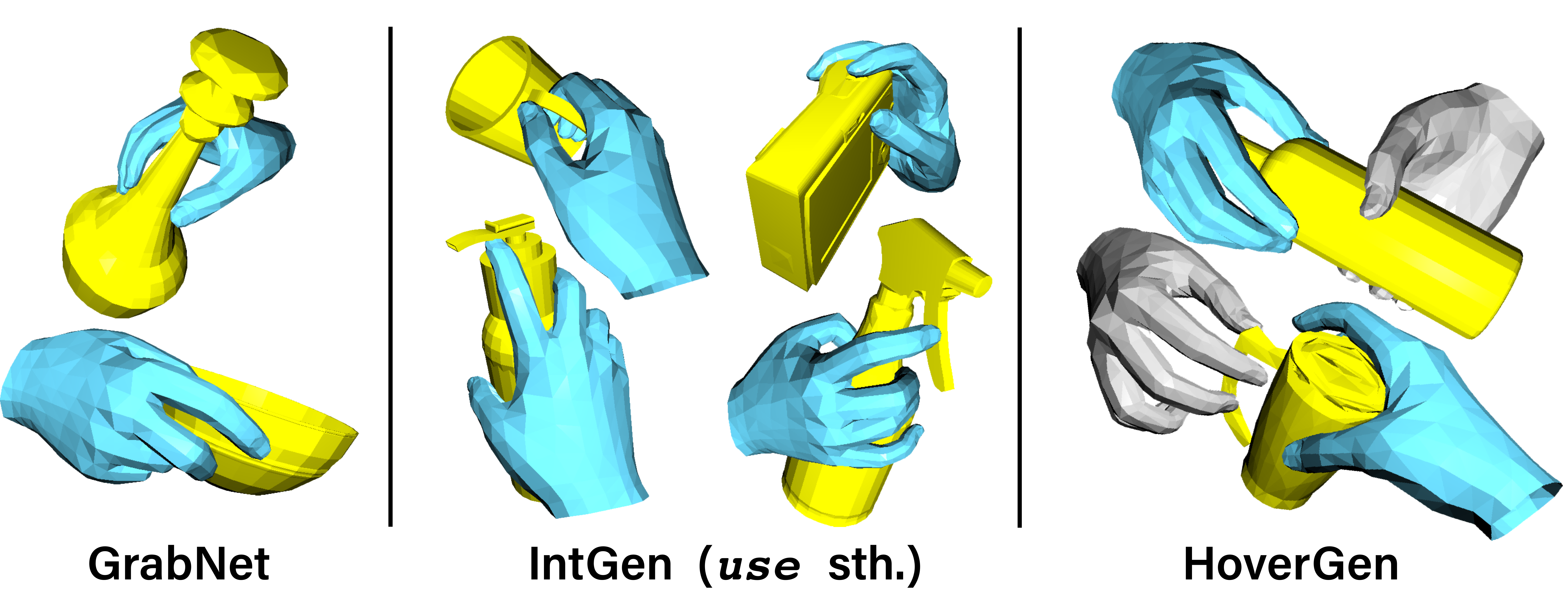}
    \end{center}\vspace{-6mm}
      \caption{Qualitative results of GrabNet, IntGen and HoverGen on \oakink-Shape. (blue: the generated hand;  gray: the giver's hand.)}\vspace{-1mm}
    \label{fig:grab_qua}
\end{figure}

\vspace{1mm}\noindent\textbf{A) Intent-Based Interaction Generation.}
We start modifying the network design in GrabNet. As shown in \cref{fig:grabnet3} \textbf{(1)+(2)},
apart from the object shape (original condition), we introduce another condition: the word embedding of a given intent. 
The model learns a hand embedding space conditioned on two dimensions: shape and intent. 
During testing, given by a test object and an assigned intent, the model decodes an intent-based interacting pose that is shown in \cref{fig:grab_qua} middle. 
We provide the evaluation results in \cref{tab:graspgen}.

\begin{table}[t]
    \centering
    \scriptsize
    \resizebox{1.0\linewidth}{!}{
        \setlength{\tabcolsep}{4.3pt}{
            \makeatletter\def\@captype{table}\makeatother
            \setlength{\tabcolsep}{1mm}{

                \begin{tabular}{l|c|cccc|c}
                    \toprule
                    \multirow{3}{*}{\textbf{Metrics} }             & \multirow{3}{*}{\tabincell{c}{\textbf{GrabNet}                                                                        \\ \cite{taheri2020grab}}}  & \multicolumn{4}{c|}{\textbf{IntGen}} & \multirow{3}{*}{\tabincell{c}{\textbf{Hover} \\ \textbf{Gen}}} \\
                    \cline{3-6}
                                                                   &                                                & \textit{mug} & \tabincell{c}{\textit{trigger}                        \\ \textit{sprayer}} & \textit{camera} & \tabincell{c}{\textit{lotion} \\ \textit{bottle}}\\
                    \midrule
                    \textit{Penet. Depth. cm}$\downarrow$          & 0.67                                           & 0.45         & 0.71                           & 1.54  & 1.57  & 0.62 \\
                    \textit{Solid Intsec. Vol. cm$^3$}$\downarrow$ & 6.60                                           & 4.22         & 9.99                           & 14.32 & 18.04 & 6.99 \\
                    \textit{Sim. Disp. Mean cm}$\downarrow$        & 1.21                                           & 0.86         & 0.69                           & 2.88  & 2.02  & 1.30 \\
                    \textit{Sim. Disp. Std cm}$\downarrow$         & 2.05                                           & 1.51         & 0.81                           & 4.53  & 2.99  & 2.03 \\
                    \textit{Percep. score} (1,..,5)$\uparrow$      & 3.66                                           & 3.86         & 3.93                           & 3.94  & 3.98  & 4.03 \\
                    \bottomrule
                \end{tabular}
            }
        }}
    \vspace{-2.5mm}
    \caption{Quantitative results on three generation tasks.} \vspace{-6.0mm}
    \label{tab:graspgen}
\end{table}

\vspace{1mm}\noindent\textbf{B) Handover Generation.}
We provide another modification of GrabNet that take the object shape as well as the giver's hand as conditions (\cref{fig:grabnet3}~\textbf{(1)+(3)}). 
The model learns to decode a receiver's hand for achieving human-to-human handover.  
We provide evaluation on several test objects in \cref{tab:graspgen} last column and \cref{fig:grab_qua} right.
The generated receiver hand matches our expectation: the receiver's hand should avoid colliding with or hindering the retraction path of the giver's hand. 

\vspace{-2mm}\section{Discussion}

\vspace{-2mm}\noindent\textbf{Limitation.} Current \oakink does not record dynamic hand interactions with the movable parts of the articulated object (\eg scissors), and does not consider transferring the interaction knowledge from human hands to the multi-finger robot arms. We will address the limitations in future works.

\vspace{1mm}\noindent\textbf{Conclusion.} In this work, we constructed a large-scale knowledge repository \oakink that builds machines' knowledge on understanding human hand-object interactions. \oakink consists of two interrelated knowledge bases \oak and \ink that contain rich data and experiences.
Even though we only benchmark \oakink on CV and CG tasks, we are quite eager to apply \oakink to the robotics community and explore future chances for robot learning. 
\noindent\rule{\columnwidth}{1.0pt}

{\small
\noindent\textbf{Acknowledgment.}~This work was supported by National Key Research and Development Project of China (No.2021ZD0110700), Shanghai Municipal Science and Technology Major Project (2021SHZDZX0102), Shanghai Qi Zhi Institute, and SHEITC (2018-RGZN-02046). 
The computing resources were provided by High-Flyer AI. 
}

{
    \small
    \bibliographystyle{config/ieee_fullname}
    \bibliography{egbib}
}

\twocolumn[{%
\renewcommand\twocolumn[1][]{#1}%
\maketitle
\begin{center}
    \centering
    \captionsetup{type=figure}
    \includegraphics[width=0.23\linewidth]{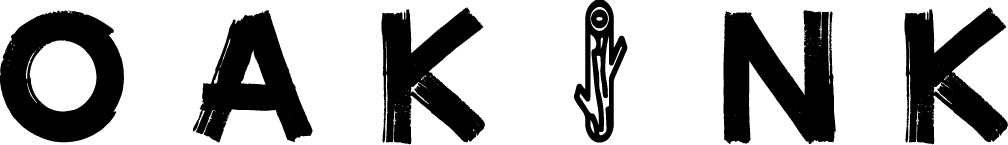}\\
    \vspace{4mm}
    {\Large\textbf{A Large-scale Knowledge Repository for Understanding}}\\
    \vspace{2mm}
    {\Large\textbf{Hand-Object Interaction}}\\
    \vspace{2mm}
    {\url{https://github.com/lixiny/OakInk}}
    \vspace{8mm}
    \label{fig:abstract}
\end{center}%
}]

\begin{appendices}\label{appendices}

\section*{\textit{Contents}}
\begin{enumerate}[font={\bfseries}, leftmargin=*]
    \setlength{\itemsep}{-0.5 mm}
    \item [\ref{sec:supp_oak_base_details}] \oak base Details;
    \item [\ref{sec:data_anno_detail}] Data Annotation Details;
    \item [\ref{sec:more_dataset_anay}] More Dataset Analysis;
    \item [\ref{sec:int_hover_gen}] Implementation: IntGen and HoverGen;
    \item [\ref{sec:perceptual}]Perceptual Survey for Generation Tasks;
    \item [\ref{sec:addi_benchmark}] Additional Benchmark Results;
    \vspace{-2mm}\begin{enumerate}[font={\bfseries}, leftmargin=7mm] 
        \item [\ref{sec:hmr}]Hand Mesh Recovery: Other Splits;
        \item [\ref{sec:outofdomain}]Unseen Out-of-domain Object;
        \item [\ref{sec:vis}]More Visualization;
    \end{enumerate}\vspace{-2mm}
    \item [\ref{sec:dis}] Discussion on Personally Identifiable Data;   
\end{enumerate}

\section{\oak Base Details}\label{sec:supp_oak_base_details}
In this section, we provide the details of the \textbf{O}bject \textbf{A}ffordance \textbf{K}nowledge base (\oak base), covering the lists of total 32 categories and 30 \textit{attribute} phrases in \cref{tab:object_taxonomy}.

\begin{table}[htp]
    \centering
    \begin{tabular}{p{50pt}p{160pt}}
    \toprule    
        { \tt \textbf{maniptool}}  & knife, screwdriver, hammer,  wrench, toothbrush, pen, frying pan, drill, pincer, scissors, stapler, mug, teapot, cup, can, box, bowl, wineglass, cylinder bottle, trigger sprayer, lotion bottle\\
    \midrule
    { \tt \textbf{functool}} & eyeglasses, headphones, binoculars, game controller, lightbulb, camera, flashlight, mouse, phone, apple, banana, donut\\
    \midrule
    \midrule
        \textbf{Attribute phrases}  &contain sth, cover sth, pump out sth, cut sth, stab sth, flow in/out sth, tighten sth, loosen sth, clamp sth, brush sth, trigger sth, 
        observe sth, point to sth, shear sth, attach to sth, connect sth, knock sth, spray sth, no function;
        hold by sth, screwed by sth, unscrewed by sth, pressed by sth, handled by sth, plug by sth, unplug by sth, squeeze by sth, pour out by sth\\
    \toprule    
    \end{tabular}\vspace{-2mm}
    \caption{The \textbf{categories} and \textbf{attribute phrases} in our \oak base}
    \label{tab:object_taxonomy}
\end{table}

\section{Data Annotation Details}\label{sec:data_anno_detail}
This section is a supplementary of the Sec. 3.2.3: \textbf{Hand Pose and Geometry}. 
Given the manually labeled 2D hand keypoints, we aim to solve the pose $\bm{\theta} \in \mathbb{R}^{16\times3}$, shape $\bm{\beta} \in \mathbb{R}^{10}$ parameters and the wrist's position $\bm{{P}}_{h,0} \in \mathbb{R}^{3}$ of a 3D hand. 
These parameters will drive a 3D hand model by a differentiable MANO layer: $\mathcal{M(\cdot)}$ \cite{romero2017embodied}:
\begin{equation}
    \bm{V}_h, \bm{P}_h = \mathcal{M}(\bm{\theta}, \bm{\beta}) + \bm{P}_{h,0}
\end{equation}
where $\bm{P}_h \in \mathbb{R}^{21\times3}$ is the hand joints' 3D position, and $\bm{V}_h \in \mathbb{R}^{778 \times 3}$ is the hand mesh vertices' 3D position.
The objective cost function for solving $\bm{\theta}$, $\bm{\beta}$ and $\bm{{P}}_{h,0}$ consists of 5 terms.

\vspace{2mm}\noindent\textbf{Reprojection Error.} 
First, we want the 2D projections of the 3D hand joints $\bm{P}_h$ to match its 2D keypoints annotation $\bm{\hat{p}}$.
Let the subscript $j$ and $v$ be the joint's ID and view's ID, we have the reprojection cost: 
\begin{equation}
    E_{\rm{repj}}=\frac{1}{\sum w_{j, v}} \sum^{4}_{v=1} \sum^{21}_{j=1} w_{j, v} \Big\| \bm{K}_v \bm{T}_v \bm{{P}}_{h,j} - \bm{\hat{p}}_{j,v} \Big\|^2_2
    \label{eq:repj_cost}
\end{equation}
The gradients from $E_{\rm{repj}}$ will back propagate to $\bm{{P}}_{h}$ and then update the $\bm{\theta}$, $\bm{\beta}$ and $\bm{{P}}_{h,0}$.

\vspace{2mm}\noindent\textbf{Geometry Consistency.}
Second, we want the 3D geometry model of hand and object to be consistent with their real-world observation: no interpenetration would occur. 
Hence, we introduce the second cost function: interpenetration loss.
We acquire the object's sign distance field: $\mathbfcal{O}$ from its scanned model, 
transform the $\mathbfcal{O}$'s pose from MoCap system to the camera system, 
and calculate the sign distance value of a 3D hand vertex $\bm{V}_{h,i}$ to $\mathbfcal{O}$.
The interpenetration cost penalizes those hand vertices inside the object surface (with negative sign distance values). 
\begin{equation}
    E_{\rm{intp}}=\sum_{\bm{V}_{h, i}} - \min \Big(\text{SDF}_{\bm{\mathbfcal{O}}}(\bm{V}_{h, i}) , 0\Big),
    \label{eq:intep_cost}\vspace{-1mm}
\end{equation}
The gradients from $E_{\rm{intp}}$ will back propagate to each $\bm{V}_{h,i}$ and then update the $\bm{\theta}$, $\bm{\beta}$ and $\bm{{P}}_{h,0}$.

\vspace{2mm}\noindent\textbf{Silhouette Constraint.}
Third, we want the contour projection of hand and object models to match the visual cues.
Hence, we introduce a binary silhouette cost.
We first acquire the hand and object's binary mask ($\mathcal{B}_h$ and $\mathcal{B}_o$) from the recorded images.
This process is automatic. 
We filter out the background pixels through green-screen and depth image matting. 
The remaining foreground pixels are the union of $\mathcal{B}_h$ and $\mathcal{B}_o$. 
Then, we render the 3D hand and object mesh on an image as silhouette and penalize the per-pixel misalignment between the rendered silhouette and the binary mask.  
\begin{equation}
  E_{\rm{sh}}=\sum_{\text{all pix.}} \underbrace{f( \mathbfcal{V}_o \cup \mathbfcal{V}_h)}_{\text{detached}}~ \cap ~BCE \Big\{ f( \mathbfcal{V}_o \cup \mathbfcal{V}_h),  (\mathcal{B}_h \cup \mathcal{B}_o) \Big\}
  \label{eq:mask_cost}
\end{equation}
In this equation, the $f(\cdot)$ is a differentiable rendering function \cite{kato2018renderer}; 
the $(\mathbfcal{V}_o \cup \mathbfcal{V}_h)$ is the composited mesh model of hand $\mathbfcal{V}_h$ and object $\mathbfcal{V}_o$;
the $f(\mathbfcal{V}_o \cup \mathbfcal{V}_h)$ is the rendered silhouette image of hand and object model;
the $(\mathcal{B}_h \cup \mathcal{B}_o)$ is the union of hand and object binary mask; 
and the $BCE \{,\}$ is the binary cross entropy loss function;
The gradients from $E_{\rm{sh}}$ will back propagate to $\mathbfcal{V}_h$ and then update the $\bm{\theta}$, $\bm{\beta}$ and $\bm{{P}}_{h,0}$.

\vspace{2mm}\noindent\textbf{Anatomical Constraint.}
Forth, we want the MANO hand pose to satisfy the anatomical constraints of human hand. 
Hence, we borrow the axial adaptations from Yang \etal \cite{yang2021cpf} and constrain the rotation axes and angles.
\begin{equation}
  E_{\rm{anat}}=\sum_{j \in \text{all}} \Big(\bm{a}_j \cdot \mathbf{n}_j^{t} + \max\big( (\phi_j - \frac{\pi}{2}), 0\big)\Big)+ \sum_{j \notin \text{MCP}} \bm{a}_j \cdot \mathbf{n}_j^{s},
  \label{eq:anat_cost}\vspace{-1mm}
\end{equation}
where the $\bm{a}_j$ and $\phi_j$ denote the axial and angular components of the $j$-th joint's rotation, the $\mathbf{n}_j^{t}$ and $\mathbf{n}_j^{s}$ are the pre-defined \textit{twist} and \textit{splay} direction, and``MCP'' indicates the five Metacarpal joints. 
The gradients from $E_{\rm{anat}}$ will back propagate to each joint's axis-angle and then update the $\bm{\theta}$. 

\vspace{2mm}\noindent\textbf{Temporal Smoothing.}
The above cost functions can only improve the per-frame precision of 3D hand annotations. 
However, frame-by-frame smoothness is also critical to improving our annotation quality.   
Hence, We want the solved 3D hand poses to be continuous in the time domain. 
We adopt a low-pass filter (\eg Kalman Filters) to post-process the poses $\bm{\theta}$ and wrist positions $\bm{{P}}_{h,_0}$ across the entire image sequence. 

\section{More Dataset Analysis}\label{sec:more_dataset_anay}

\begin{figure}[h]
  \begin{center}
    \includegraphics[width=0.98\linewidth]{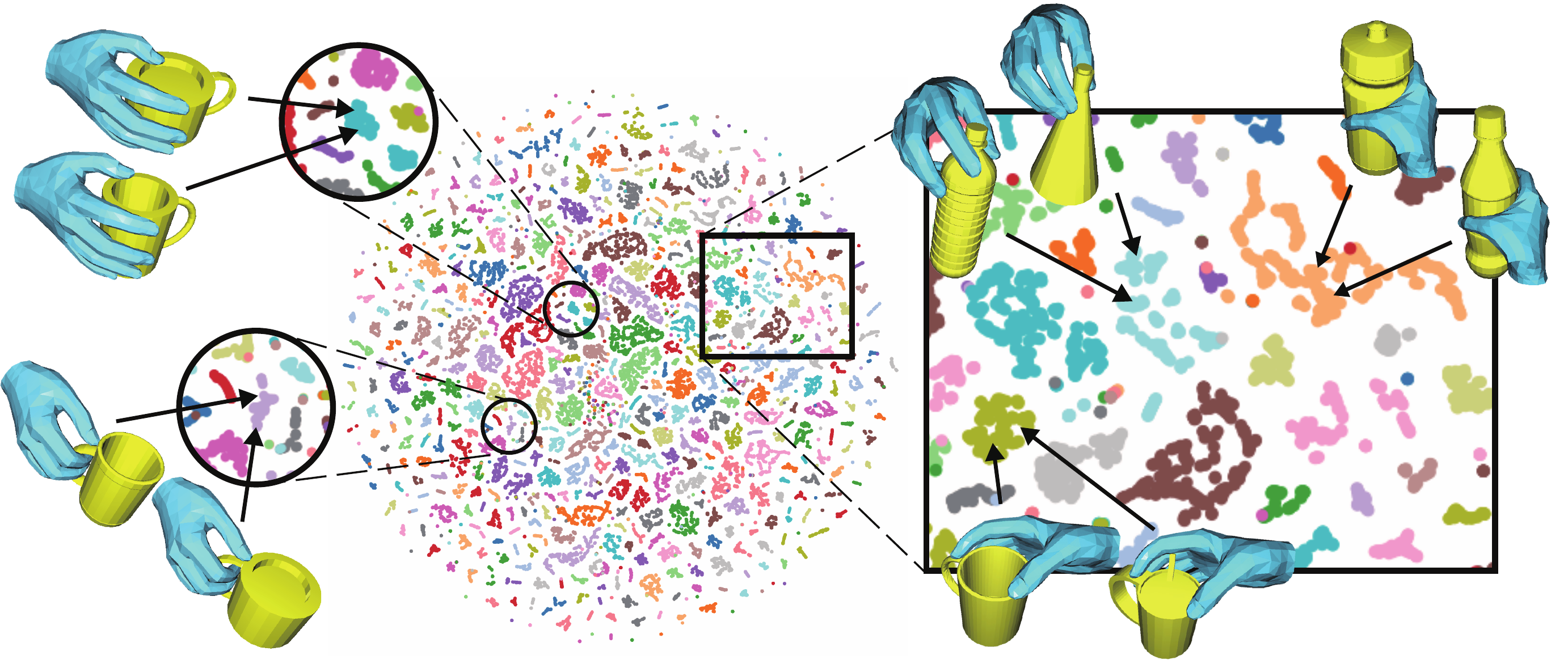}
  \end{center}\vspace{-4mm}
  \caption{t-SNE embedding of hand poses. We randomly select 20 colors to visualize the clustered poses. }\vspace{-2mm}
  \label{fig:pose_distr}
\end{figure}

\vspace{1 mm}\noindent\textbf{Hand Pose Distribution.}
We project the interacting hand poses into an embedded space yield from t-SNE \cite{t-SNE}. The poses that transferred from the same \oakink-Core pose are painted in the same color.
From the box in \cref{fig:pose_distr}, we can see that the similar interacting hand poses with different objects are mapped to adjacent in the embedded 2D space.
From the circles in \cref{fig:pose_distr}, we can conclude that the different grasping types are away from each other.

\begin{figure}[h]
  \begin{center}
    \includegraphics[width=0.9\linewidth]{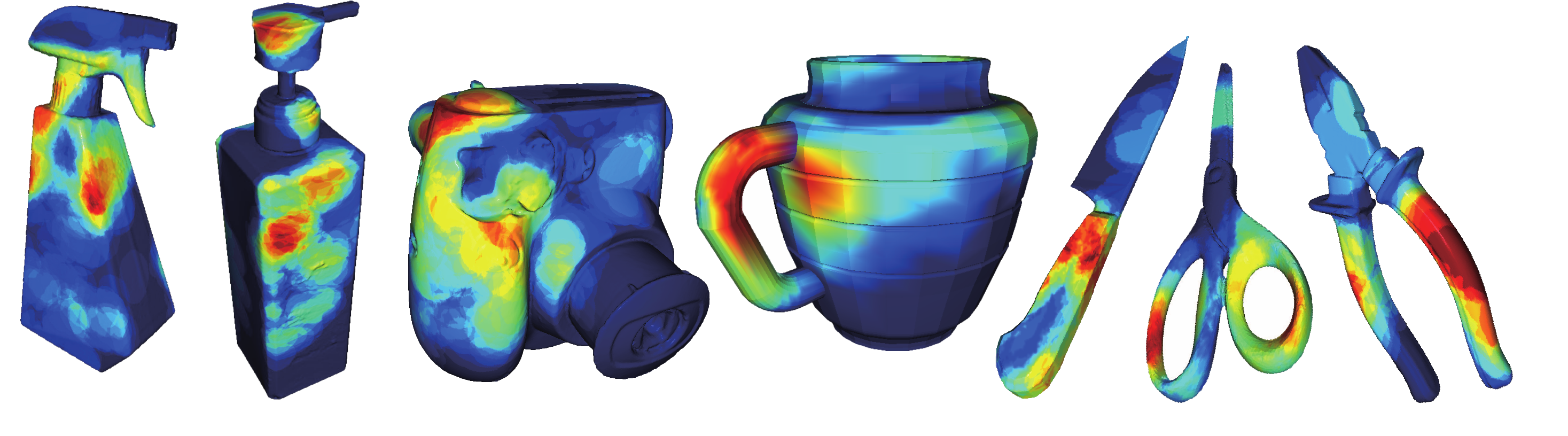}
  \end{center}\vspace{-4mm}
  \caption{Heatmaps of contact frequency on object surface.}
  \label{fig:contact_distr}\vspace{-2mm}
\end{figure}

\vspace{1 mm}\noindent\textbf{Contact Distribution.}
We provide the contact heatmaps on example objects that reveal the frequencies of contact among all interactions.
\cref{fig:contact_distr} shows such heatmaps on six \oak base categories.
We see that the ``hot'' area (red) that denotes the high frequency of contact is consistent with the object affordance we described.

\section{Implementation: IntGen and HoverGen}\label{sec:int_hover_gen}
\begin{figure}[h]
    \begin{center}
        \includegraphics[width=0.98\linewidth]{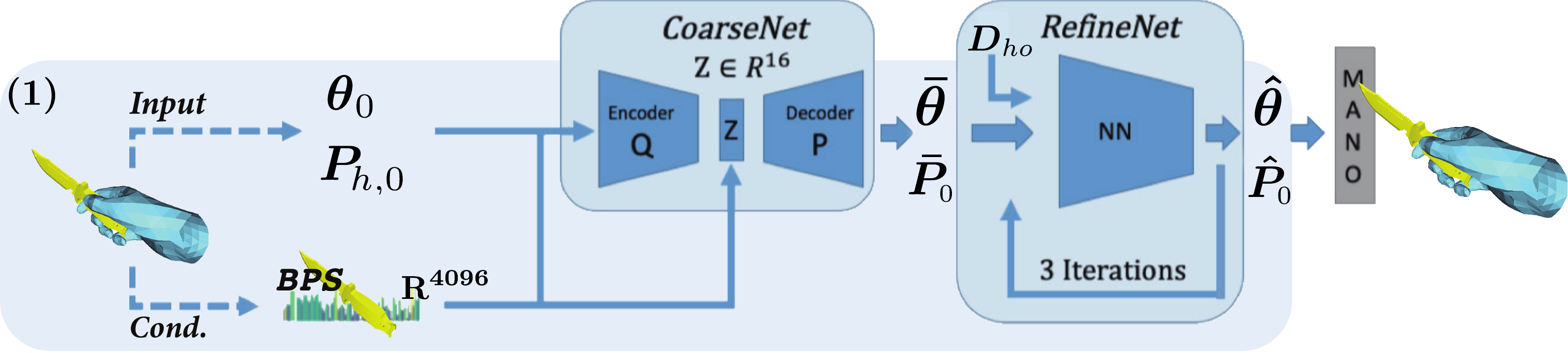}
    \end{center}\vspace{-4mm}
        \caption{\textbf{GrabNet}\cite{taheri2020grab}: the original design.}
    \label{fig:grab}
\end{figure}

\begin{figure}[h]
    \begin{center}
        \includegraphics[width=0.98\linewidth]{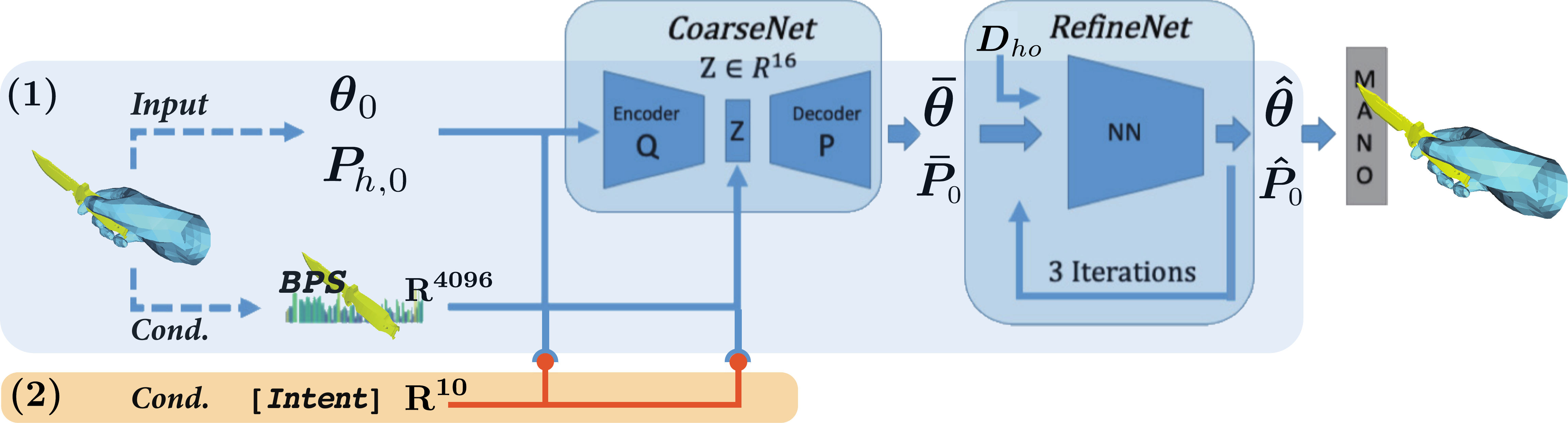}
    \end{center}\vspace{-4mm}
        \caption{\textbf{IntGen}: the intent-based grasp generation network design.}
    \label{fig:intgen}
\end{figure}

\begin{figure}[h]
    \begin{center}
        \includegraphics[width=1.0\linewidth]{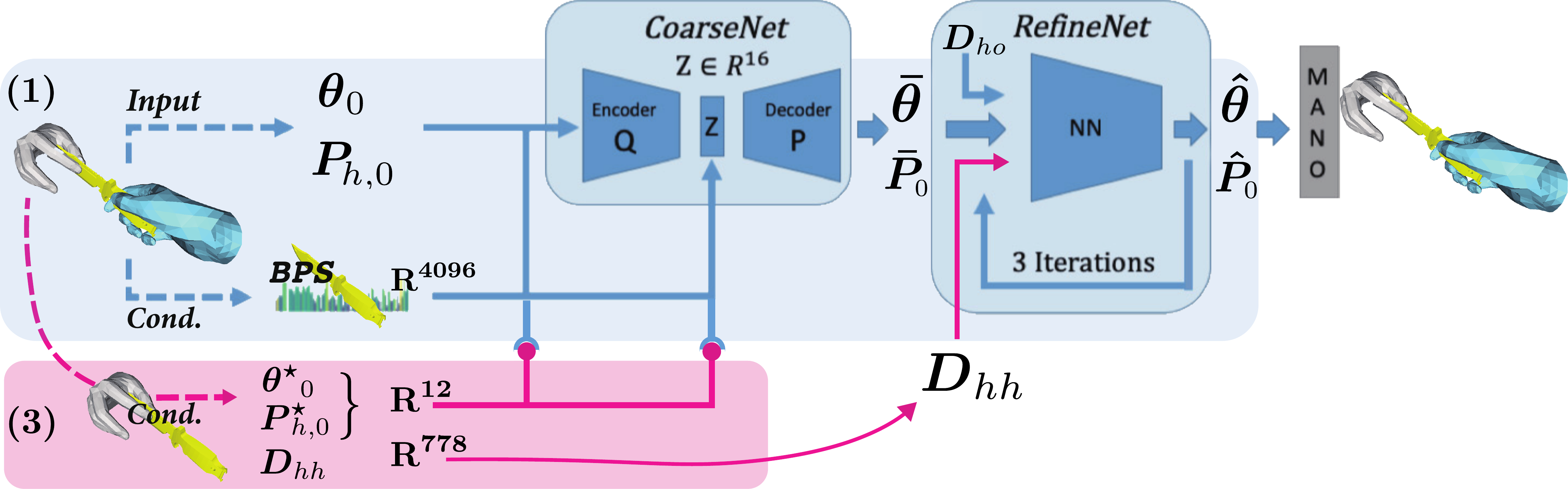}
    \end{center}\vspace{-4mm}
        \caption{\textbf{HoverGen}: the handover generation network design. The giver's hand is paint in \textbf{\color[RGB]{120, 120, 120}gray} and the receiver's hand in \textbf{\color[RGB]{102, 190, 243}blue}. }\vspace{-2mm}
    \label{fig:hovergen}
\end{figure}

The architecture of IntGen (\cref{fig:intgen}) and HoverGen  (\cref{fig:hovergen}) model are modified from the original GrabNet \cite{taheri2020grab} (\cref{fig:grab}) design.

In the IntGen task, we select three intents: \textit{use}, \textit{hold} and \textit{hand-out}, map the intents' word string to a real-valued word vector, and train the networks with the intent vector as the additional input.
During training, poses within different intents will be mapped to different areas in the latent pose space: $\mathcal{Z} \in \mathbb{R}^{16}$. The training loss functions in IntGen are identical to those in GrabNet, including standard conditional VAE losses (KL divergence and weight regularization), mesh reconstruction losses  (hand vertices and mesh edge loss), and physical quality losses (penetration and contact loss). 
We train the IntGen on category-level data in \oakink-Shape, including the \textit{mug}, \textit{camera}, \textit{trigger sprayer} and \textit{lotion bottle}. The training process lasts 1,000 epochs, with the mini batch size 32 and initial learning rate of $1\times10^{-3}$. The learning rate decays by a factor of 0.5 at every 200 epochs. 

As shown in \cref{fig:hovergen}, in the HoverGen task, we provide the root rotation: $\bm{\theta^{\star}}_0$ and root position: $\bm{P^{\star}}_{h,0}$ of the giver's hand as the additional inputs for CoarseNet, 
and the Chamfer distance: $\bm{D_{hh}} \in \mathbb{R}^{778}$ from the original giver's hand to the predicted receiver's hand as an additional input for RefineNet. As a result, the HoverGen model learns a receiving hand's embedding space, $\mathcal{Z}$, conditioned on the object shape and the giver's hand root pose. At inference time, given an unseen object shape and the giver's hand root pose: ($\bm{\theta^{\star}}_0, \bm{P^{\star}}_{h,0}$), we sample a vector from $\mathcal{Z}$ and decode a receiver hand pose to complete a human-to-human handover. 
The training loss in HoverGen model includes all the losses in IntGen model, 
plus an L1 loss on the
Chamfer distance $\bm{D_{hh}}$ \wrt the ground-truth $\bm{\hat{D}_{hh}}$, a Chamfer distance from the original giver's hand to the ground-truth receiver's hand.
We train the HoverGen model 1,000 epochs with a mini-batch size 256 and an initial learning rate of $1\times10^{-3}$, decaying a half at every 200 epochs.

\section{Perceptual Survey for Generation Tasks}\label{sec:perceptual}
\begin{figure*}[htb]
    \begin{center}
        \includegraphics[width=1.0\linewidth]{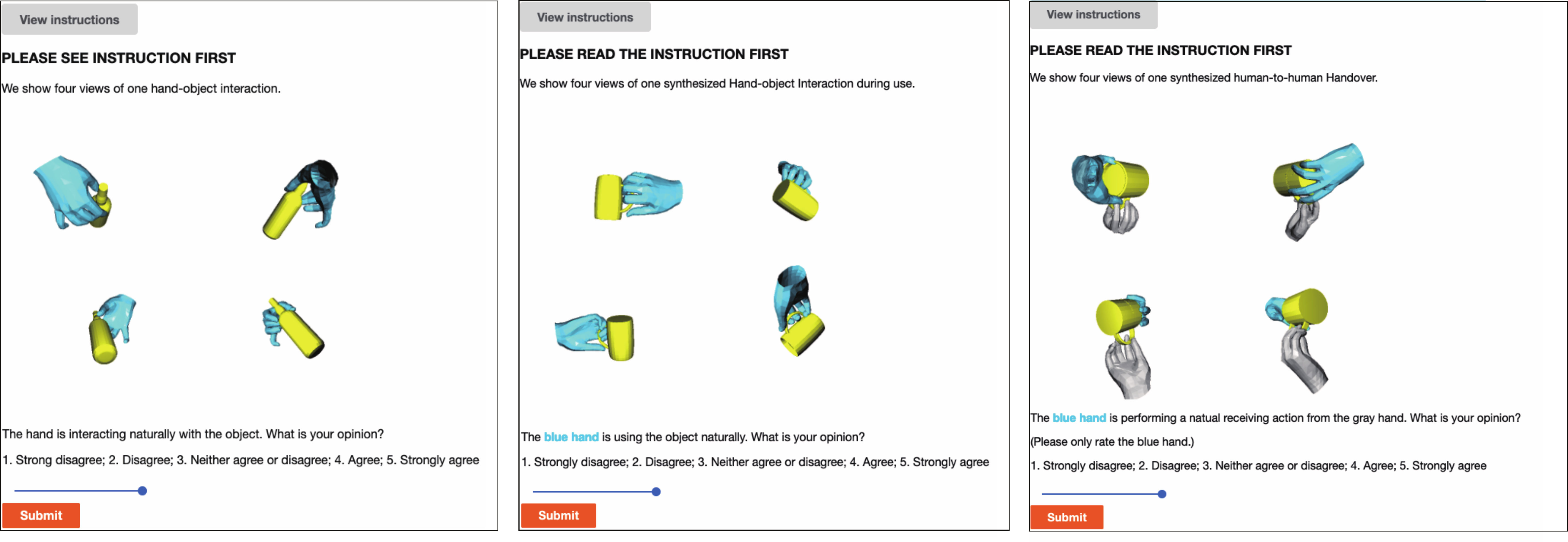}
    \end{center}\vspace{-3mm}
        \caption{\textbf{The layout of three perceptual surveys on AMT}.\\ Left: GrabNet (statement: \textit{the hand is interacting naturally with the object}); \\Middle: IntGen (statement: \textit{the blue hand is using the object naturally}); \\Right: HoverGen (statement: \textit{the blue hand is performing a natural receiving action from the gray hand})}\vspace{-1mm}
    \label{fig:amt}
\end{figure*}

\vspace{-1mm}To investigate the general audience's opinion about the predicted pose of the generation tasks: GrabNet, IntGen, and HoverGen, we conducted three perceptual surveys on the Amazon Mechanical Turk (AMT). 
In each survey, we show four views of each predicted hand-object interaction and ask the audiences to give their opinion about a statement (\eg ``\textit{the hand is interacting naturally with the object}''). The audiences are asked to rate the statement with a 5-level Likert scale (``strongly agree'' corresponds to grade 5 and ``strongly disagree'' corresponds to grade 1). 
The layout of the perceptual surveys on GrabNet, IntGen, and HoverGen are shown in \cref{fig:amt}.

\section{Additional Benchmark Results}\label{sec:addi_benchmark}
\subsection{Hand Mesh Recovery: Other Splits}\label{sec:hmr}
Apart from the default split {\tt \textbf{SP0}} (split by views) in the main text, we also provide another two data splits and the HMR benchmark results for \oakink-Image. 
\begin{itemize}[font={\bfseries}, leftmargin=*]
    \item {\tt \textbf{SP1}} \textbf{(subjects split).} (train/val/test: 6/1/5). We split the \oakink-Img by subjects. The subjects recorded in the test split will not appear in the train split. 
    \item {\tt \textbf{SP2}} \textbf{(objects split).} (train/test: 70\%/5\%/25\%). We split the \oakink-Img by objects. The objects that have been grasped in the test split will not appear in the train split. 
\end{itemize}

\begin{table}[h]
\centering
\resizebox{0.9\linewidth}{!}{
\footnotesize
\setlength{\tabcolsep}{4.3pt}
    \begin{tabular}{l|l|c|c}
        \toprule
                        \textbf{Splits}  & \textbf{Methods} & \textbf{MPJPE}$\downarrow$ (AUC$\uparrow$) &  \textbf{MPVPE}$\downarrow$ \\
        \midrule
        \multirow{2}{*}{\tt \textbf{SP1}} & I2L-MeshNet~\cite{moon2020i2lmehsnet}   & 18.04  (\textit{0.641}) & 18.08   \\
                                            & HandTailor~\cite{lv2021handtailor}  & 15.72  (\textit{0.792})& 16.31 \\
        \midrule
        \multirow{2}{*}{\tt \textbf{SP2}} & I2L-MeshNet~\cite{moon2020i2lmehsnet}   & 15.79  (\textit{0.733}) & 15.87   \\
                                            & HandTailor~\cite{lv2021handtailor}  & 14.14  (\textit{0.846}) & 14.81 \\
        \bottomrule
    \end{tabular}
}
\caption{ \textbf{HMR results in \textit{mm}}. AUC are shown in parentheses.}
\label{tab:hmr}
\end{table}

\subsection{Unseen Out-of-domain Object}\label{sec:outofdomain}
\begin{figure}[h]
    \begin{center}
        \includegraphics[width=1.0\linewidth]{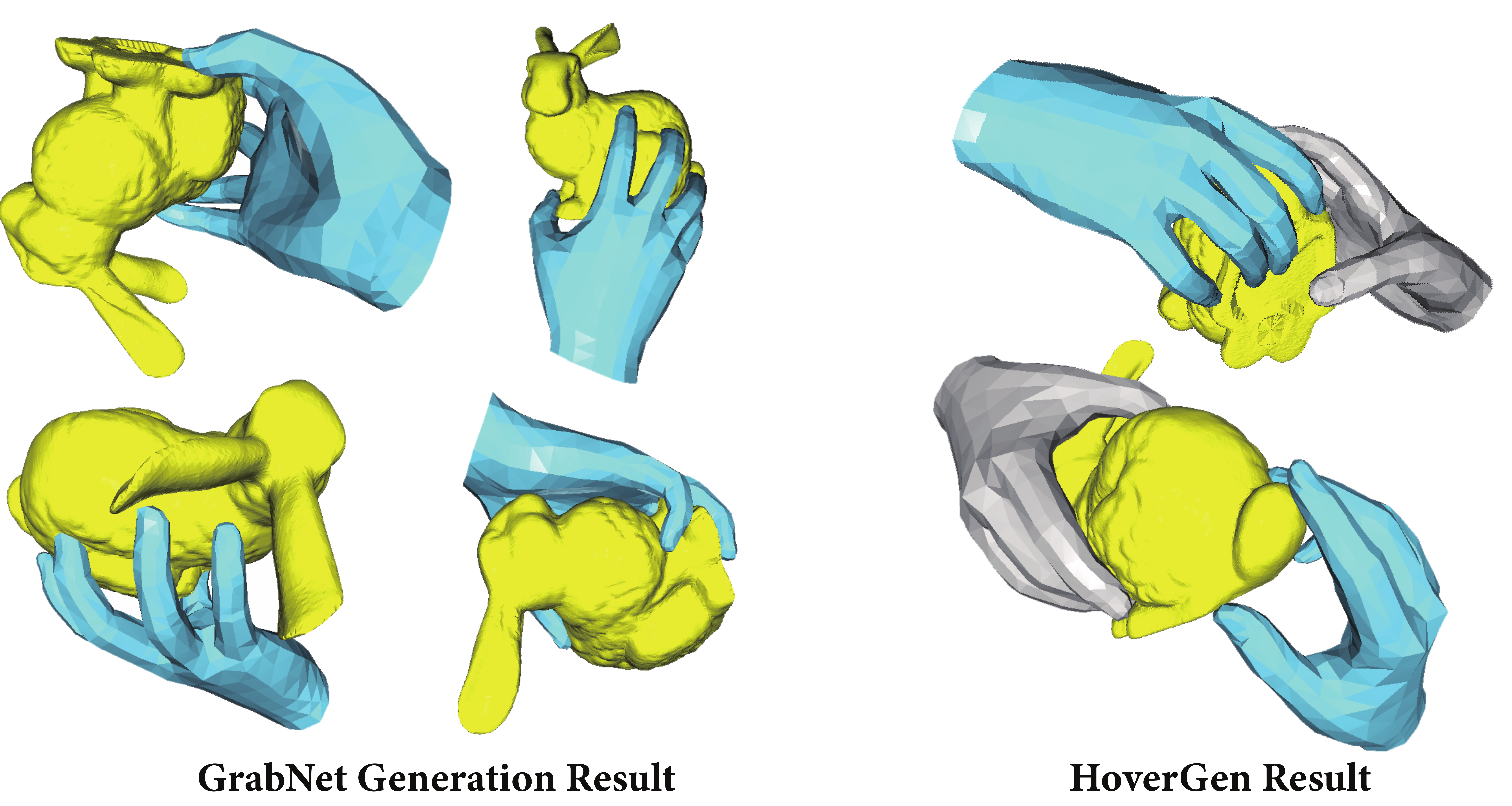}
    \end{center}
        \caption{Generation results on \textbf{unseen out-of-domain} objects.}
    \label{fig:ood}
\end{figure}

We refer the objects in \oakink-Shape test set as unseen in-domain objects, indicating that they may have similar counterparts included in the training set. 
In this part, we are also interested in the performance of our generation tasks on the unseen \textbf{out-of-domain} objects. 
We choose the Stanford bunny, a general 3D test model, as an illustrative prototype of \textbf{out-of-domain} object. 
We test the GrabNet and HoverGen model on the Stanford bunny and provide the generated grasps and receiving poses in \cref{fig:ood}.
Both the GrabNet and HoverGen model are trained on our \oakink-Shape training set. 
The results show that through training on the \oakink-Shape, GrabNet and HoverGen can synthesize realistic and prehensile interactions for general objects.  

\subsection{More Visualization}\label{sec:vis}
\begin{figure*}[htb]
    \begin{center}
        \includegraphics[width=0.97\linewidth]{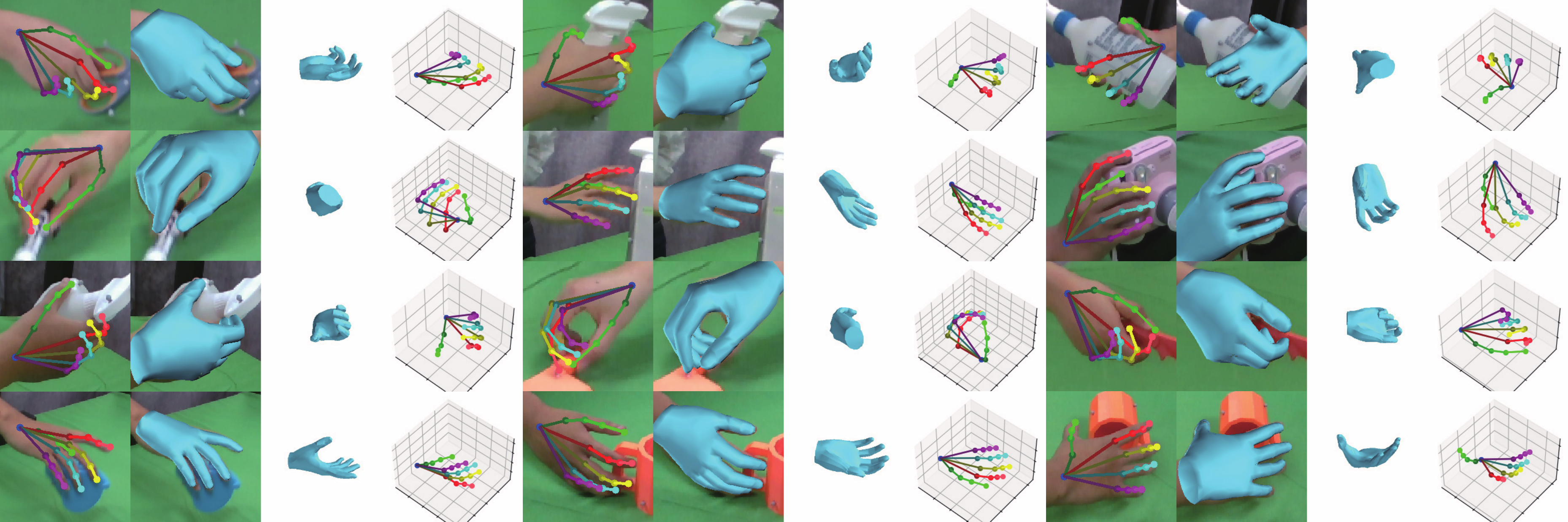}
    \end{center}
        \caption{\textbf{More qualitative results} on HMR task.}
    \label{fig:visual_hmr}
\end{figure*}

\begin{figure*}[htb]
    \begin{center}
        \includegraphics[width=0.97\linewidth]{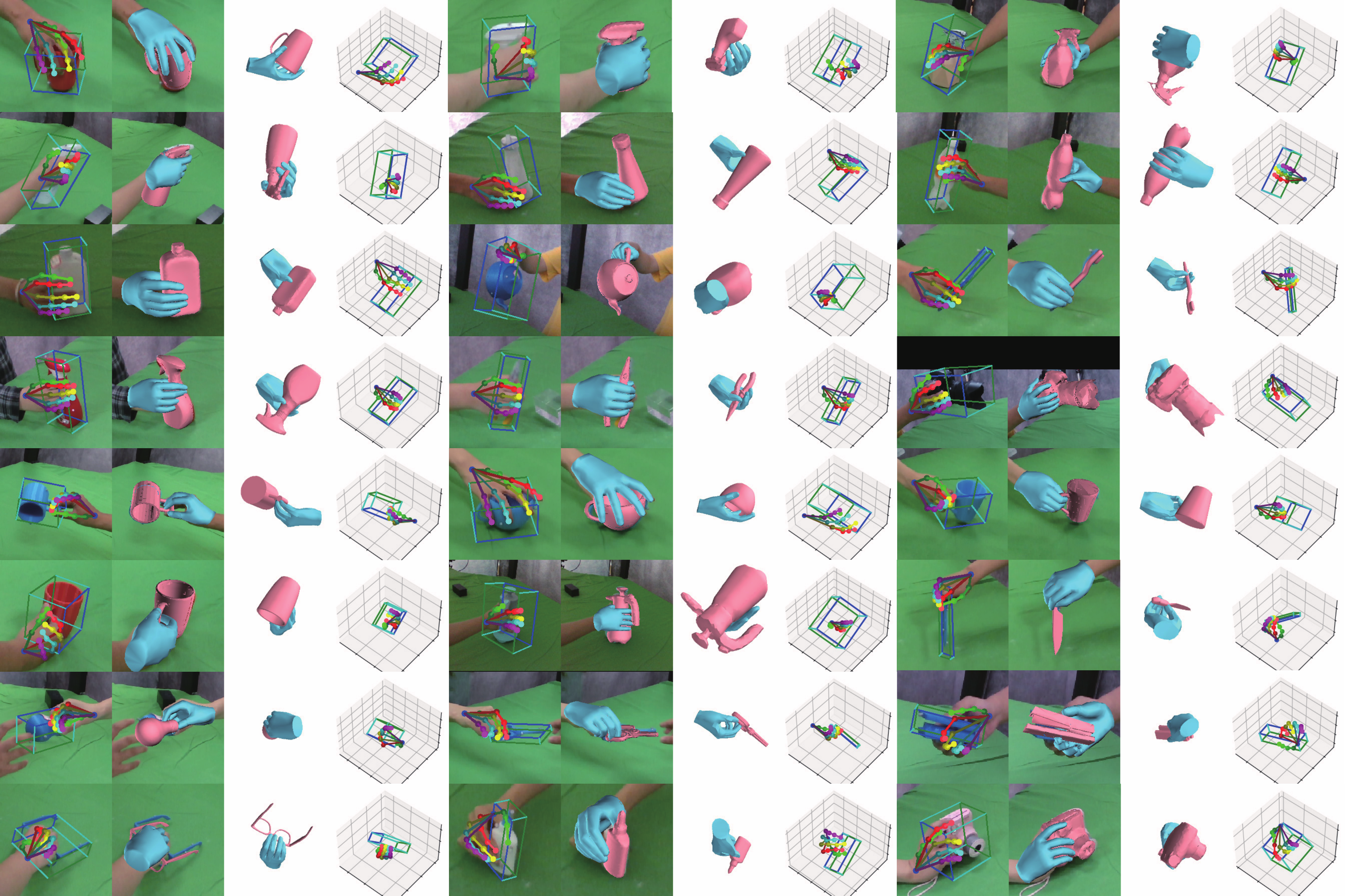}
    \end{center}
        \caption{\textbf{More qualitative results} on HOPE task.}
    \label{fig:visual_hope}
\end{figure*}

\begin{figure*}[htb]
    \begin{center}
        \includegraphics[width=0.97\linewidth]{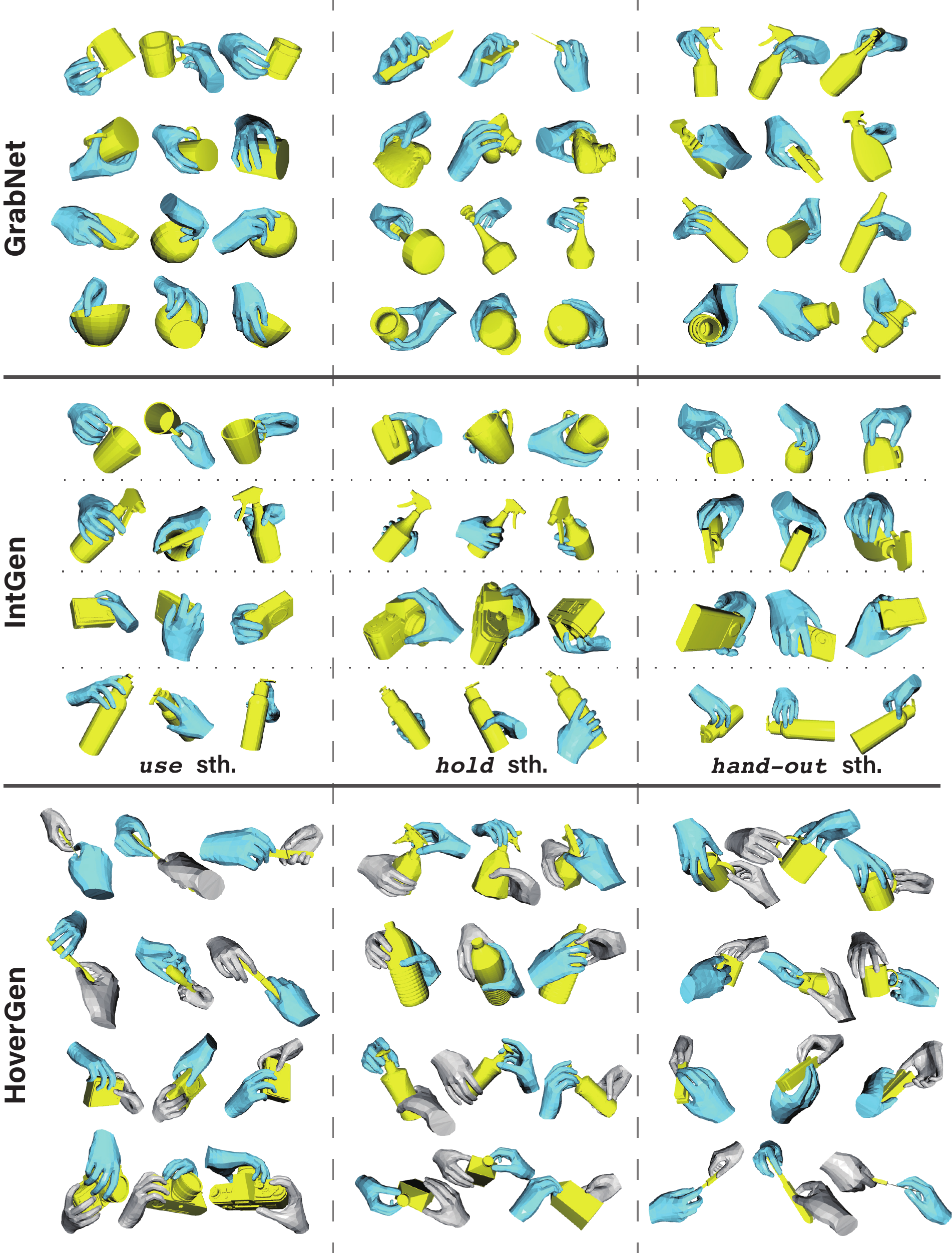}
    \end{center}\vspace{-6mm}
        \caption{\textbf{More qualitative results} on GrabNet, IntGen and HoverGen predictions.}
    \label{fig:visual_gen}
\end{figure*}

We provide more qualitative results of the benchmark results of HMR task in \cref{fig:visual_hmr}, HOPE task in \cref{fig:visual_hope}, GraspGen (GrabNet) in \cref{fig:visual_gen}: top, IntGen in \cref{fig:visual_gen}: middle, and HoverGen in \cref{fig:visual_gen}: bottom.

\section{Discussion on Personally Identifiable Data}\label{sec:dis}
We collect hand-object interaction data on 12 human subjects recruited through a third-party crowd-sourcing company. In the collection process, their actions will be recorded in video sequences by the MulCam system. We ensure that the data collecting process meets the ethics requirements through the following announcements:

\vspace{-2mm}\begin{itemize}[font={\bfseries}, leftmargin=*]
\setlength{\itemsep}{-1mm}

\item The third-party crowd-sourcing company warrants appropriate IRB approval (or equivalent, based on local government requirements) are obtained. 
The company name and warranties are withheld based on the anonymous submission guidelines.

\item All the subjects involved in data collection are required to sign a contract with the third-party crowd-sourcing company, involving permission on the portrait usage, the acknowledgment of data usage, and payment policy. 
During the data collecting process, all subjects are paid by the hour.

\item All the subjects involved in the data collecting process acknowledge that the collected data will only be intended for academic and permitted commercial usages.

\item We ensure all the subjects involved in the data collecting process are willing to share the personal-related data, including actions, skin tones, body/hand shapes, \etc.

\item We require all the subjects not to dress in revealing or offensive clothes during the data collection process.

\item Upon the release of the dataset, we will desensitize all samples in the dataset by blurring the subjects' faces (if any), tattoos, rings, or any other accessories that may reveal the subjects' identity.

\end{itemize}

\clearpage

\end{appendices}

\end{document}